\documentclass{article}
\usepackage{arxiv}
\usepackage[utf8]{inputenc} % allow utf-8 input
\usepackage[T1]{fontenc}    % use 8-bit T1 fonts
\usepackage{hyperref}       % hyperlinks
\usepackage{url}            % simple URL typesetting
\usepackage{booktabs}       % professional-quality tables
\usepackage{amsfonts}       % blackboard math symbols
\usepackage{amsmath}
\usepackage{amssymb}
\usepackage{nicefrac}       % compact symbols for 1/2, etc.
\usepackage{microtype}      % microtypography
\usepackage{multirow}
\usepackage{graphicx}
\usepackage{natbib}
\usepackage{doi}

\title{Single-shot Star-convex Polygon-based Instance Segmentation for Spatially-correlated Biomedical Objects}

\author{ \href{https://orcid.org/0000-0003-1111-9851}{\includegraphics[scale=0.06]{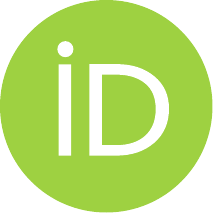}\hspace{1mm}Trina De} \\
	Department of Computer Science\\
    Technical University of Dresden \\
    Dresden, Germany \\
	Helmholtz-Zentrum Dresden-Rossendorf e. V.\\ 
	Bautzner Landstrasse 400, 01328 Dresden \\
	\texttt{t.de@hzdr.de} \\
	\And
	\href{https://orcid.org/0009-0008-6619-3665}{\includegraphics[scale=0.06]{orcid.pdf}\hspace{1mm}Adrian Urba\a'nski} \\
	Institute of Computer Science\\
    University of Wrocław\\
	Wrocław, Poland \\
    Helmholtz-Zentrum Dresden-Rossendorf e. V.\\ 
	Bautzner Landstrasse 400, 01328 Dresden \\
	\texttt{a.urbanski@hzdr.de} \\
    \And
	\href{https://orcid.org/0000-0003-2458-4904}{\includegraphics[scale=0.06]{orcid.pdf}\hspace{1mm}Artur Yakimovich}\thanks{Corresponding author is Artur Yakimovich (a.yakimovich@hzdr.de).}\\
	Institute of Computer Science\\
    University of Wrocław\\
	Wrocław, Poland \\
    Helmholtz-Zentrum Dresden-Rossendorf e. V.\\ 
	Bautzner Landstrasse 400, 01328 Dresden \\
	\texttt{a.yakimovich@hzdr.de} \\
}

\renewcommand{\headeright}{}
\renewcommand{\undertitle}{}

\hypersetup{
pdftitle={Single-shot Star-convex Polygon-based Instance Segmentation for Spatially-correlated Biomedical Objects},
pdfauthor={Trina De, Adrian Urba\a'nski, Artur Yakimovich},
pdfkeywords={biomedical imaging, branched architectures, instance segmentation, spatial correlation, star-convex polygons},
}

\begin{document}
\maketitle

\begin{abstract}
	Biomedical images often contain objects known to be spatially correlated or nested due to their inherent properties, leading to semantic relations. Examples include cell nuclei being nested within eukaryotic cells and colonies growing exclusively within their culture dishes. While these semantic relations bear key importance, detection tasks are often formulated independently, requiring multi-shot analysis pipelines. Importantly, spatial correlation could constitute a fundamental prior facilitating learning of more meaningful representations for tasks like instance segmentation. This knowledge has, thus far, not been utilised by the biomedical computer vision community. We argue that the instance segmentation of two or more categories of objects can be achieved in parallel. We achieve this via two architectures \emph{HydraStarDist (HSD)} and the novel (\emph{HSD-WBR}) based on the widely-used \emph{StarDist (SD)}, to take advantage of the star-convexity of our target objects. \emph{HSD} and \emph{HSD-WBR} are constructed to be capable of incorporating their interactions as constraints into account. HSD implicitly incorporates spatial correlation priors based on object interaction through a joint encoder. \emph{HSD-WBR} further enforces the prior in a regularisation layer with the penalty we proposed named \emph{Within Boundary Regularisation Penalty (WBR)}. Both architectures achieve nested instance segmentation in a single shot. We demonstrate their competitiveness based on \emph{$IoU_R$} and \emph{AP} and superiority in a new, task-relevant criteria, \emph{Joint TP rate (JTPR)} compared to their baseline SD and Cellpose. Our approach can be further modified to capture partial-inclusion/-exclusion in multi-object interactions in fluorescent or brightfield microscopy or digital imaging. Finally, our strategy suggests gains by making this learning single-shot and computationally efficient.
\end{abstract}


\section{Introduction}
\label{sec:intro}
Analysis of biomedical images often deals with objects that are strongly spatially-correlated or even nested. While such strong relationships of objects may not always hold true in natural scenes, in the case of microscopy, pathology and tomography, it is expected that the co-location of organs and organelles is generally conserved and has a semantic meaning. Nuclei are to be found within the eukaryotic cells, and observable phenotypes typically occur within the respective assay dishes. This notion is, however, rarely taken advantage of in the widespread biomedical datasets or deep learning (DL) architectures.

To address this, here we introduce a formulation of a penalty for DL architectures for biomedical object segmentation that facilitates the segmentation of spatially-correlated objects in the biomedical domain in a single shot. We use a modified \textbf{H}ydra\textbf{S}tar\textbf{D}ist (\emph{HSD}) architecture and propose a new \textbf{H}ydra\textbf{S}tar\textbf{D}ist with \textbf{W}ithin \textbf{B}oundary \textbf{R}egularisation (\emph{HSD-WBR}) and compare the performance of these architectures to \textbf{S}tar\textbf{D}ist (\emph{SD}) \citep{Stardist18} and \textbf{C}ell\textbf{p}ose  (\emph{CP}) \citep{Cellpose21} - the state-of-the-art (SOTA) architectures for cell and nuclei segmentation. Both HSD and HSD-WBR integrate the knowledge of spatial correlation implicitly by using a joint encoder in a modified SD architecture for the two different yet nested objects. Furthermore, in \emph{HSD-WBR} we attempt to solve the problem of false predictions of the inner nested object outside the outer nested object (e.g. predictions of nuclei outside cells or plaques outside the tissue culture plates) explicitly. Specifically, \emph{HSD-WBR} further enforces the prior of spatial correlation using the Within Boundary Regularisation penalty.

We thus propose a unified branched framework, which incorporates a modifiable penalty for representing multi-object correlations. The results reveal that our method is competitive based on traditionally known criteria for the instance segmentation task and outperforms SD and Cellpose based on our new proposed single-shot segmentation-relevant criterion \emph{JTPR} (see Section \ref{sec:methods} and Tables \ref{tab:joint_tp_rate_hela_branched_foundational_performance}-\ref{tab:joint_tp_rate_vacv_branched_foundational_performance}). They are also lighter on compute than SD and requires significantly lesser training time. Using our methods we are able to achieve single-shot two-channel instance segmentation for the nested objects in contrast to the setup of two different experiments.

To demonstrate the benefits of this approach we explored two datasets HeLaCytoNuc (Subsection \ref{subsection:hela_dataset}) and VACVPlaque (Subsection \ref{subsection:plaque_dataset}). HeLaCytoNuc (Subsection \ref{subsection:hela_dataset}) is an open annotated dataset of fluorescence micrographs with spatially-correlated cytoplasm and nuclei instances. In healthy eukaryotic cells, nuclei are generally expected to be within the cytoplasm, suggesting a nested relationship. Due to the spatial uniformity and relatively good separation, stained nuclei are often selected as the primary object to detect in microscopy and digital pathology \citep{Cellprofiler06, Qupath21}. The detection of cytoplasm, in turn, often represents a greater challenge due to the significant overlap and rather irregular shape. In the case of VACVPlaque (Subsection \ref{subsection:plaque_dataset}), the spatially-correlated objects are the virological plaques – circular phenotypes of vaccinia virus (VACV, a prototypic poxvirus \citep{Vaccinia09}) spread, and wells of the assay plate.

\begin{table}[h!]
    \caption{HeLaCytoNuc Overall Performance. \emph{JTPR} (Equations \ref{eq:jtp_rate_inner}-\ref{eq:jtp_rate_outer}) with respect to the inner and outer nested objects, parameter count and training time on HeLaCytoNuc (Subsection \ref{subsection:hela_dataset}). \textbf{Best}. \underline{Second best}.}
  \centering
  \resizebox{0.75\linewidth}{!}{
  \begin{tabular}{ccccc}
    \toprule
    & inner nested & outer nested & & \\
    Architecture & object $\uparrow$ & object $\uparrow$ & \#Parameters $\downarrow$ & Training $\downarrow$ \\
    & (\emph{Nuclei}) & (\emph{Cytoplasm}) & ($\approx$ millions) & (mins) \\
    \midrule
    \emph{HSD-WBR} (ours) & \textbf{0.898} & \textbf{0.855} & \underline{30.2} &  \underline{207} \\
    \emph{HSD} (ours) & \underline{0.880} & \underline{0.837} & \underline{30.2} & \textbf{206} \\
    \emph{SD} (RI) \citep{Stardist18} & 0.817 & 0.771 & 44.4 & 277 \\
    \emph{SD} (FT) \citep{Stardist18} & 0.814 & 0.768 & 44.4 & 231 \\
    \emph{CP} (RI) \citep{Cellpose21} & 0.820 & 0.773 & $\mathbf{13.2}$ & 280 \\
    \emph{CP} (FT) \citep{Cellpose21} & 0.776 & 0.730 & $\mathbf{13.2}$ & 240 \\
  \bottomrule
  \end{tabular}
  }
  \label{tab:joint_tp_rate_hela_branched_foundational_performance}
\end{table}

\begin{table}[h!]
  \caption{VACVPlaque Overall Performance. \emph{JTPR} (Equations \ref{eq:jtp_rate_inner_vacv}-\ref{eq:jtp_rate_outer_vacv}) with respect to the inner and outer nested , parameter count and training time on VACVPlaque (Subsection \ref{subsection:plaque_dataset}). Experiments with Cellpose were omitted due to to its applicability only to cellular objects. \textbf{Best}. \underline{Second best}.}
  \centering
  \resizebox{0.75\linewidth}{!}{
  \begin{tabular}{ccccc}
    \toprule 
    & inner nested & outer nested & & \\
    Architecture & object $\uparrow$ & object $\uparrow$ & \#Parameters $\downarrow$ & Training $\downarrow$ \\
    & (\emph{Plaques}) & (\emph{Wells}) & ($\approx$ millions) &  (mins) \\
    \midrule
    \emph{HSD-WBR} (ours) & \underline{0.784} & \textbf{0.849} & \textbf{30.2} & \underline{17} \\
    \emph{HSD} (ours) & 0.755 & 0.833 & \textbf{30.2} & \underline{17} \\
    \emph{SD} (RI) \citep{Stardist18} & \textbf{0.833} & \underline{0.841} & \underline{44.4} & 26 \\
    \emph{SD} (FT) \citep{Stardist18} & 0.746 & \textbf{0.849} & \underline{44.4} & \textbf{16} \\
    \bottomrule
  \end{tabular}
  }
  
  \label{tab:joint_tp_rate_vacv_branched_foundational_performance}
\end{table}

\section{Related Work}
\label{sec:related_works}
\subsection{Branched models in the Biomedical domain}
\label{subsection:branched_instance_segmentation}
\begin{figure}[!t]
    \centering
    \includegraphics[width=0.5\columnwidth]{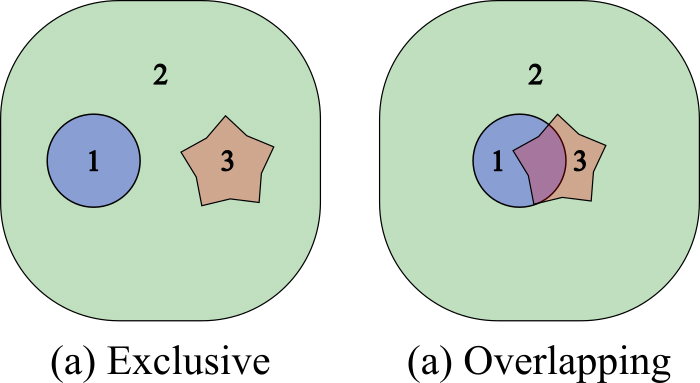}
    \caption{Limitations of approach in \citep{Delong09}. Here object 2 bounds objects 1 and 3 that may either be (a) Exclusive or (b) Partially Overlapping with each other. $\alpha$ is the fraction of overlap between 1 and 3 (Equation \ref{eq:wbr_loss_limitation_overlap}).}
    \label{fig:limitation}
\end{figure}

Branched architectures have long been used to take advantage of parallel streams of information and make for an efficient training setup when the tasks are related. A branched architecture based on U-Net \citep{Unet15} modified for object detection and subsequent segmentation for occluded transparent overlapping objects of a single class can be found in \citep{Isoodl18, Isoodlv219}. However, their work focuses on solving the problem of occlusion (faced by U-Net) within the same object category by converting 2D masks into 3D by a shearing technique along the z-axis. Our task focuses on fully nested objects that we try to predict simultaneously in one end-to-end model. Our approach is also different from \citep{Branched123}, which uses a multi-branch synchronous learning segmentation network based on local and global information to finally achieve better segmentation results on a single class of objects. \citep{Branched223} is closely related to the approach in \citep{Branched123} but the task isn't in the biomedical domain. An important body of work is on capturing the object part relations. Delong and Boykov in \citep{Delong09} propose a way of representing object part relations through graph cut optimisation. Specifically, \citep{Delong09, Delong10, Sonka06} seem relevant to the domain and task, while \citep{Zhao18} splits identified human regions into sections that do not necessarily include interacting object parts. An important caveat of \citep{Delong09} is its limitation of representing multiobject co-locations such as when an outer object type contains two interior located object types that are exclusive to or partially overlapping with each other (Figure \ref{fig:limitation}). In
\ref{sec:methods}, we show that our formulation of the constraint is able to express such a relationship.

\begin{figure}[!t]
    \centering
    \includegraphics[width=0.75\columnwidth]{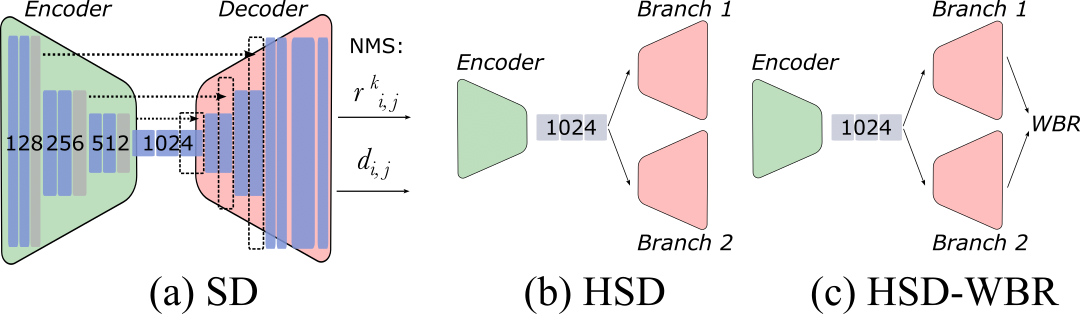}
    \caption{StarDist-based instance segmentation architectures 
    (Section \ref{sec:methods}). Architecture diagrams of the instance segmentation model for star-convex objects (a) StarDist, (b) branched architecture based on the SD model
    HydraStarDist and (c) branched architecture with WBR penalty HydraStarDistWBR (Section \ref{sec:methods}).}
    \label{fig:architectures}
\end{figure}

\subsection{Instance Segmentation Metrics}
\label{subsection:instance_segmentation_metrics}
Average Precision (AP) (Equation \ref{eq:ap}) \citep{Iou1912} is a standard evaluation metric for instance segmentation (identifying each individual relevant object uniquely). It is also the performance metric reported in SOTA models StarDist and Cellpose. However, due to its insufficiency in capturing the nature of the task and its nature, such as non-differentiability and decomposability, modifications have been suggested to it in the works \citep{Mao19,Ramzi21}. \citep{Chen23} states for AP-based metrics that they are only conditionally sensitive to the change of masks or false predictions. For certain metrics, the score can change drastically in a narrow range, which could provide a misleading indication of the quality gap between results. In Section \ref{sec:methods}, we show a case where a change in \emph{true positives (TP)} and a related change in \emph{false positives (FP)} would leave the value of AP unchanged. We also introduce a set of new evaluation criterion, \emph{JTPR} variants (Equations \ref{eq:jtp_rate_inner}-\ref{eq:jtp_rate_outer_vacv}) which captures better the nested instance segmentation task. And finally, in this paper, we base our conclusions about performance on \emph{JTPR} and \emph{$IoU_R$}
(Equation \ref{eq:iou-recall}) (which can be thought of as a version of Recall \citep{Recall1955}, wherein the \textit{IoU} between the ground truth mask and prediction is assigned as the score for every TP instead of 1 in the numerator in the standand Recall formula) rather than AP but report those values too keeping in line with StarDist \citep{Stardist18} and Cellpose \citep{Cellpose21}.

\subsection{Plaque and Nuclei Quantification}
\label{subsection:plaque_and_nuclei_quantification}
Machine Learning (ML)-based approaches to biomedical objects, especially plaque quantification, have been proposed in the past \citep{Cacciabue19,Phanomchoeng22}. But keeping the ML algorithms employed relatively simplistic comes at the cost of laborious and standardised image acquisition techniques (flatbed scanner or an apparatus assembly), which is avoidable as our approach shows. In contrast, we use SD \citep{Stardist18}-based CNNs \citep{Lecun15}, that are widely used, leveraging star-convex polygon-based modeling and enabling the harnessing of digital-photographic information. We argue that the flexibility of our methods and proven results on microscopy data will further facilitate the development of advanced algorithms for virological plaque, cell, nuclei and general nested biomedical object quantification.

\section{Methods}
\label{sec:methods}

\subsection{Star-convexity}
\label{subsection:star-convexity}
Our baseline and modifications are based on the modelling of biomedical objects as star-convex polygons, as defined below.

\textbf{Definition 1} (Star-convex sets) \citep{Topology00}. \emph{Let V be a vector space over $\mathbb{R}$ or $\mathbb{C}$. A subset A $\subseteq$ V is said to be \textbf{star-convex} if and only if $\ \exists$ a $\in$ A such that:}
\begin{equation}
\centering
\label{eq:star-convexity}
    \forall x \in A : \forall t \in [0,1] : tx + (1-t)a \in A
\end{equation}
The point \emph{a $\in$ A} is called a \textbf{star centre} of \emph{A}. A \textbf{star-convex set} can thus be described as a set containing all line segments between the star centre and an element of the set.
We can also have \emph{star-convex functions} where every linear combination of functions in a function class is once again a member of the class \citep{Starconvex16}. In our case we have a set A $\subseteq V = \mathbb{R}^2$. 

\subsection{StarDist Principles}
\label{subsection:stardist}
Using the idea of star-convexity, StarDist \citep{Stardist18} tries to approximate an object using a star-convex polygon of \emph{K} vertices. For each pixel \emph{(i,j)} StarDist predicts two values. The first is \textbf{Boundary distance probabilities}: $d_{i,j}$, the normalised Euclidean distance to the nearest background pixel. Post normalisation, these can be treated as and act like foreground/background probabilities depending on their proximity to the perphery. When non-maximal suppression (NMS) \citep{NMS1986} is applied, it favours polygons associated with pixels near the object centre, which typically represents objects more accurately. The second is \textbf{Star-convex polygon radial distances}: For every pixel belonging to an object, $r^k_{i,j}$ is defined as the Euclidean distance to the object boundary along $k$, each of which is one of \emph{K} equispaced radial directions around $2\pi$, $k \in \{0, 1, 2, ..., K-1\}$. Using a combination of the above two quantities, they construct star-convex polygon proposals that are further thresholded using NMS \citep{NMS1986}.

A set of individual ground truth radial distances ${r}^k_{i,j}$ are generated for each individual pixel $(i,j)$ based on the instance mask. Star-convex polygon proposals are made for all the pixels, and NMS \citep{NMS1986} is applied to filter overlapping candidates by taking into account their associated boundary distance $d_{i,j}$ as probabilities. The final set of polygons not rejected by NMS, outline the object instances in the image.

\subsection{Losses}
\label{subsection:losses}
For the baseline task of single-channel instance segmentation we use the architecture SD (Figure \ref{fig:architectures}), where to predict star-convex polygons the loss is based on a combination of $r^k_{i,j}$
and $d_{i,j}$. Namely, the \textbf{Object Boundary Loss} ($\mathcal{L}_{OBL}$ based on the normalised Euclidean distance to the nearest background pixel $d_{i,j}$) and \textbf{Distance Loss} ($\mathcal{L} _{DL}$ based on $r^k_{i,j}$, the Euclidean distance to the nearest background pixel along $k$, one of \emph{K} equi-spaced radial directions around $2\pi$, $k \in \{0, 1, 2, ..., K-1\}$).The total loss $\mathcal{L}$ is defined as,
\begin{equation}
\centering
    \label{eq:baseline_loss_abbreviated}
    \mathcal{L} = \mathcal{L}_{OBL} + \lambda_1 \cdot \mathcal{L}_{DL}
\end{equation}

where $\mathcal{L}_{OBL}$ measures the error in $d_{i,j}$ between the predicted and the ground truth. It uses a Binary Cross-Entropy Loss (BCE) \citep{Bce1948} to enforce accurate boundary prediction.

The $\mathcal{L}_{DL}$ quantifies the error in predicting $r^k_{i,j}$ along the \emph{K} equi-spaced radial directions. In this case a Mean Absolute Error (MAE) \citep{Mae2022} or $L_1$ Loss term is used to minimise a regression loss, between predicted and ground truth distances. $\lambda_1$ serves as a regularisation factor, that regulates over or under-prediction of polygon vertices distances from the star centre \citep{Topology00,Starconvex16} along radial directions $k$.
\newline
As written and explained concretely in \citep{Splinedist21}, the baseline loss can effectively be written as,
\begin{equation}
\centering
\label{eq:baseline_loss}
    \begin{split}
    \mathcal{L} = & \mathcal{L}_{BCE}(d_{i,j}, \hat{d}_{i,j})\ + \\
                  & \lambda_1 \cdot (d_{i,j} \cdot 1_{d_{i,j}>0} \cdot \frac{1}{K}\sum_{k=0}^{K-1} |r^k_{i,j} - \hat{r}^k_{i,j}| + \\
                  & \lambda_2 \cdot 1_{d_{i,j}=0} \cdot \frac{1}{K}\sum_{k=0}^{K-1} |\hat{r}^k_{i,j}|)\\
    \end{split}
\end{equation}

 $\lambda_2$ regularises and prevents polygon predictions around star centres (Equation \ref{eq:star-convexity}) where $d_{i,j}$ is 0, i.e. polygon predictions with background pixels as the centre in the ground truth.This value of this loss is computed by averaging its value over all the pixels in the image.

The total loss over all pixels in our proposed architecture HSD-WBR is defined as,
\begin{equation}
\centering
\label{eq:combined_loss}
    \mathcal{L}^{\prime} = \sum_{i,j} \mathcal{L}_1 + \sum_{i,j} \mathcal{L}_2 + \lambda_3 \cdot \Lambda
\end{equation}

where $\mathcal{L}_1$ and $\mathcal{L}_2$ are individually the same as $\mathcal{L}$ in Equation \ref{eq:baseline_loss} for each pixel for the two branched decoders respectively (Figure \ref{fig:architectures}). $\lambda_3$ is a regularisation factor that helps to penalise nested object predictions that violate our prior knowledge criteria and $\Lambda$ is the WBR penalty defined as,
\begin{equation}
\centering
\label{eq:wbr_loss}
    \Lambda = \Big(1+\epsilon - \cfrac{\sum_{i,j} (\hat{y}^{2 \perp}_{i,j} * \hat{y}^{1}_{i,j})}{\sum_{i,j} y^{2 \perp}_{i,j}}\Big)^{-1}
\end{equation}
where $\epsilon > 0$ is an arbitrarily small value. $\hat{y}^{2 \perp}_{i,j}$ is the $(i,j)th$ pixel of the inverted predicted semantic mask (obtained from predicted instance mask) of the outer nested object i.e. cytoplasm or wells. $\hat{y}^{1}_{i,j}$ is the  $(i,j)th$ pixel of the predicted semantic mask (obtained from for predicted instance mask) of the inner nested object i.e. nuclei or plaque. ($*$) represents element-wise product. \\
The value of the $\Lambda$ therefore lies in the bounded open interval $(\frac{1}{1+\epsilon}, \frac{1}{\epsilon})$, lower limit being when there are no predictions of inner nested object outside the outer nested object and upper limit when every pixel outside the boundaries of the predicted outer nested object is a prediction of the inner nested object. It is not a problem that the lower limit is not 0 since in the best case we are only adding an upper-bounded constant term to the total loss which is $\frac{1}{1+\epsilon} \rightarrow 1$ as $\epsilon \rightarrow 0$. The upper limit to the WBR penalty is an open interval since in the worst case, $\Lambda = \frac{1}{\epsilon} \rightarrow +\infty$ as $\epsilon \rightarrow 0$ which for arbitrarily small values of $\epsilon$ can be infinitely large. A very large value here works in our favour to enforce strongly the regularisation.\\

Here, we penalise the prediction of inner nested object instances ${y}^{1}$ outside the boundary of outer nested objects ${y}^{2}$ (which according to prior knowledge should not occur). Implicitly, we also penalise errors in the prediction of the total and individual area of outer object instances ${y}^{2}$ i.e. cytoplasm or wells. Specifically, the under-prediction of this outer object area, since an under-prediction would mean that $\sum_{i,j} (\hat{y}^{2 \perp}_{i,j} * \hat{y}^{1}_{i,j})$ would be high, leading to a higher $\Lambda$. \\
Our architecture HSD does not implement the WBR penalty and therefore the loss here is simply $\mathcal{L}^{\prime}$ from Equation \ref{eq:combined_loss} without the $\Lambda$ term. \\

Additionally, we are able to achieve more flexibility in representing relationships between nested objects than the formulation in \citep{Delong09} using a modification of the penalty term $\Lambda$ in Equation \ref{eq:wbr_loss}. For better understanding, please read the below segment with Figure \ref{fig:limitation} in mind. In case the two interior objects are exclusive to one another, we can write the penalty as, 
\begin{equation}
\centering
\label{eq:wbr_loss_limitation_exclusive}
    \begin{split}
    \Lambda^{'} &= \Big(3 +\epsilon - \mathcal{L}_{1-interior} - \mathcal{L}_{3-interior} - \mathcal{L}_{13-exclusion} \Big)^{-1} \\
    &\mathcal{L}_{1-interior} = \cfrac{\sum_{i,j} (\hat{y}^{2 \perp}_{i,j} * \hat{y}^{1}_{i,j})}{\sum_{i,j} y^{2 \perp}_{i,j}} \\
    &\mathcal{L}_{3-interior} = \cfrac{\sum_{i,j} (\hat{y}^{2 \perp}_{i,j} * \hat{y}^{3}_{i,j})}{\sum_{i,j} y^{2 \perp}_{i,j}} \\
    &\mathcal{L}_{13-exclusion} = \cfrac{\sum_{i,j} (\hat{y}^{1}_{i,j} * \hat{y}^{3}_{i,j})}{\sum_{i,j} y^{1}_{i,j}}
    \end{split}
\end{equation}

In case of partial overlap between the inner nested objects, the penalty takes the form,
\begin{equation}
\centering
\label{eq:wbr_loss_limitation_overlap}
    \begin{split}
    \Lambda^{''} = \Big(2 + max(\alpha,1-\alpha)^{2} + \epsilon - \mathcal{L}_{1-interior} - \mathcal{L}_{3-interior} - \\
    &\mathcal{L}_{13-overlap} \Big)^{-1} \\
    \mathcal{L}_{13-overlap} = \Big(\alpha - \cfrac{\sum_{i,j} (\hat{y}^{1}_{i,j} * \hat{y}^{3}_{i,j})}{\sum_{i,j} y^{1}_{i,j}}\Big)^{2}
    \end{split}
\end{equation}

where $0 < \alpha < 1$ indicating the fraction of expected overlap and $\mathcal{L}_{1-interior}$, $\mathcal{L}_{3-interior}$ are as in Equation \ref{eq:wbr_loss_limitation_exclusive}.

The value of the $\Lambda^{'}$ and $\Lambda^{''}$ therefore lies in the bounded open interval $(\frac{1}{3+\epsilon}, \frac{1}{\epsilon})$ (second term in the denominator of $\Lambda^{''}$ is so written since the range of $\mathcal{L}_{13-overlap}$ is $(0, max(\alpha,1-\alpha)^{2})$, lower limit being when all the conditions of the relationship are satisfied in the predictions and upper limit being when none of them are satisfied. The values can once again be arbitrarily large depending on $\epsilon$.

As also explained in Section \ref{sec:intro}, StarDist is biomedical objects seen in microscopy, and therefore, widely applicable including digital photography data. Cellpose, on the other hand, is only applicable (validated for our datasets) (See Section \ref{sec:experiments}) to cellular objects seen in microscopy. Since we do not build upon or modify in any way the losses that Cellpose employs, we do not analyse them here in detail and refer readers instead to the excellent original Cellpose paper \citep{Cellpose21} for details.

\subsection{Evaluation Criteria}
\label{subsection:metrics}
We have reported performance on the nested instance segmentation task on two metrics from \citep{Stardist18}, Intersection over Union over recalled objects ($IoU_R$) and Average Precision ($AP$) and our proposed \emph{Joint TP rate (JTPR)} that is relevant specifically to this task with emphasis on $IoU_R$ and \emph{JTPR}. \\
\newline
$IoU_R$ (\href{https://github.com/stardist}{StarDist repository}) is defined as,
\begin{equation}
\centering
\label{eq:iou-recall}
    IoU_R(\tau) = \sum_{o_i, \hat{o}_i \forall i \in (O \cap \hat{O})_\tau} \frac{IoU(o_i, \hat{o}_i)}{|O|}
\end{equation}

where $IoU$ is defined in \citep{Iou1912}, and $O$ and $\hat{O}$ is the set of all ground truth objects and predicted objects respectively. $|O|$ is the number of count of ground truth objects. Therefore, $(O \cap \hat{O})_\tau$ is the set of all ground truth objects predicted correctly at \emph{IoU} \citep{Iou1912} threshold $\tau \in (0,1)$. Equation \ref{eq:iou-recall} can be imagined as Recall \citep{Recall1955} with a score of $IoU(o_i, \hat{o}_i)$ per correctly recalled object (TP) instead of 1 in the numerator. Similar to Recall \citep{Recall1955} where we normalise by $TP+FN$, we normalise here with the equivalent of $TP+FN$ i.e. $|O|$.\\
\newline
$AP$ (\href{https://github.com/stardist}{StarDist repository}) is defined as,
\begin{equation}
\centering
\label{eq:ap}
  AP(\tau) = \frac{TP_\tau}{TP_\tau + FN_\tau + FP_\tau}
\end{equation}
based on two sets $O$ and $\hat{O}$ as defined above.\\
$TP_\tau = |\{ (o_{i},\hat{o}_{i}) \in (O \cap \hat{O}) | IoU(o_i, \hat{o}_i) > \tau\}|$, $\tau \in (0,1)$. \\
Similarly we have,\\
$FN_\tau = |\{ (o_{i},\hat{o}_{i}) | (o_{i} \in O, \hat{o}_i \not\in \hat{O})\ or\ (o_{i} \in O, \hat{o}_i \in \hat{O}, IoU(o_i, \hat{o}_i) < \tau) \}|$ and\\
$FP_\tau = |\{ \hat{o}_{i} \in (\hat{O} \setminus O) \}|$. $\setminus$ is  the set difference operator \citep{Set1980}. $TN_\tau = |\{(o_{i},\hat{o}_{i}), o_{i} \not\in O, \hat{o}_{i} \not\in \hat{O}\}|$ is infeasible to report since the count of this can be combinatorially infinite (the number of incorrect star-convex polygons that are not in the ground truth). \\
The caveat of AP is as follows. There could be a undesired change in FP and TP that leaves the value of AP unchanged or greater which is a poorer task performance but is not manifested in the metric. From Equation \ref{eq:ap}, let,
\begin{equation}
\centering
\label{eq:ap_limitation}
  AP(\tau) = \frac{TP_\tau}{TP_\tau + FN_\tau + FP_\tau} = \frac{a}{a+b+c}
\end{equation}
where $a > 0, b, c \ge 0, a+b = constant(k)$ since $TP + FN$ is a constant. Let's say, between two instance segmentation results, $TP$ decreases, then $\Delta TP < 0$ and $\Delta FN = - \Delta TP$ by definition.  Now we need the condition such that there is no effect of the change of TP and FP on AP. We need not consider separately, the effect of change of FN since jointly with TP it is a constant for an image. So we try to find this condition by solving the below inequality (especially where this bound is tight),
\begin{equation}
\centering
\label{eq:ap_limitation_relation}
    \begin{split}
    &\frac{a - \Delta a}{k + c - \Delta c} \ge \frac{a}{k+c} \\
    &\implies  \frac{(a - \Delta a)(k + c)}{a} \ge k + c - \Delta c \\
    &\implies \frac{(a - \Delta a)(k + c)}{a} - (k+c) \ge - \Delta c \\
    &\implies \Delta c \ge  (k+c) - \frac{(a - \Delta a)(k + c)}{a} \\
    &\implies \Delta c \ge \frac{ak + ac - ak - ac - \Delta a(k+c)}{a} \\
    &\implies \Delta c \ge - \frac{\Delta a (k+c)}{a} \\
    &\implies \Delta c \ge  - \frac{\Delta a(k+c)}{a} > 0 \because \Delta a \le 0 \\
    \end{split}   
\end{equation}
which implies that for a poorer model showing an $\Delta FP > 0$ and $\Delta TP < 0$, AP could still remain the same or increase when $\Delta FP = -\frac{\Delta TP(k+FP)}{TP}$ or greater when $LHS > RHS$. $IoU_R$ (Equation \ref{eq:iou-recall}) does not suffer from the same drawback (by definition). Both however do not take into account the spatial-correlation of our nested objects.\\
\newline
Therefore we propose our own evaluation criteria, \textit{Joint TP rate} (JTPR) (w.r.t. the inner nested object) which is defined as,
\begin{equation}
\centering
\label{eq:jtp_rate_inner}
    JTPR_{inner} = \cfrac{
    |({O}_{2} \cap \hat{O}_{2}) \cap ({O}_{1} \cap \hat{O}_{1})|}{| O_{1}|}
\end{equation}
where ${O}_{1}$ and $\hat{O}_{1}$ represent the true inner nested objects and their predictions. Similarly, ${O}_{2}$ and $\hat{O}_{2}$ represent the true outer nested objects and their predictions. Since ${O}_{1}$ could be unequal to ${O}_{2}$ \emph{i.e.} not all outer nested objects have inner nested counterparts or vice versa, we also report another variation, \textit{JTPR} (w.r.t. the outer nested object),
\begin{equation}
\centering
\label{eq:jtp_rate_outer}
    JTPR_{outer} = \cfrac{ 
    |({O}_{1} \cap \hat{O}_{1}) \cap ({O}_{2} \cap \hat{O}_{2})|}{| O_{2}|}
\end{equation}
Equations \ref{eq:jtp_rate_inner}-\ref{eq:jtp_rate_outer} arrive at the same value in their numerators set-theoretically, and are more apt at capturing the nature of our task here since it is sure to show a decreases when TP falls for either of the object categories.\\
We show yet another variation where the one outer nested object contains multiple inner nested objects as is the case shown in Figure \ref{fig:limitation} and in one of our datasets, VACVPlaque (Subsection \ref{subsection:plaque_dataset}), used for experiments. In this case $JTP\ rate_{inner}$ and $JTP\ rate_{outer}$ are defined as,
\begin{equation}
\centering
\label{eq:jtp_rate_inner_vacv}
    JTPR_{inner} = \cfrac{
    \sum_{\substack{o_{j} \in \\ ({O}_{2} \cap \hat{O}_{2})}}|({O}_{2} \cap \hat{O}_{2}) \cap ({O}_{1j} \cap \hat{O}_{1j})|}{|O_{1}|}
\end{equation}
and \\
\begin{equation}
\centering
\label{eq:jtp_rate_outer_vacv}
    JTPR_{outer} = \cfrac{
    \sum_{\substack{o_{j} \in \\ ({O}_{2} \cap \hat{O}_{2})}}|({O}_{1j} \cap \hat{O}_{1j}) \cap ({O}_{2} \cap \hat{O}_{2})|}{| O_{2}|}
\end{equation}
respectively, where ${O}_{1j}$ and $\hat{O}_{1j}$ represent the true and predicted inner nested objects corresponding to the $j^{th}$ member of $({O}_{2} \cap \hat{O}_{2})$. Just like the flexible WBR penalty representation (Subsection \ref{subsection:losses}), \textit{JTPR} can also be modified by verying depending on multiobject interactions, the set membership rules of the individual components above. Here we have represented 1-1 total inclusions (Equations \ref{eq:jtp_rate_inner}-\ref{eq:jtp_rate_outer}) and 1-many total inclusions (Equations \ref{eq:jtp_rate_inner_vacv}-\ref{eq:jtp_rate_outer}) only.

All metric values are averaged over the test images and are reported for the optimal $\tau$ for $IoU_R$ and $AP$ in Tables \ref{tab:joint_tp_rate_hela_branched_foundational_performance}-\ref{tab:vacv_foundational_performance}.

\section{Experiments}
\label{sec:experiments}

\subsection{HeLaCytoNuc Dataset}
\label{subsection:hela_dataset}

\begin{figure}[!t]
    \centering
    \includegraphics[width=0.75\columnwidth]{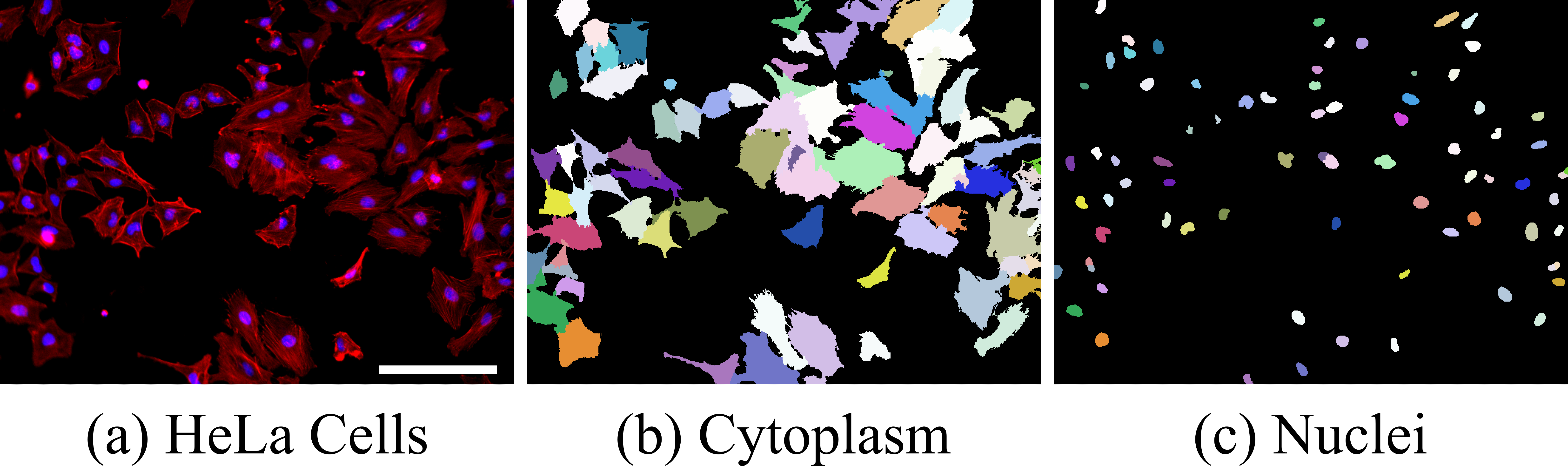}
    \caption{HeLaCytoNuc \citep{rodare_HeLaCytoNuc24} dataset (Subsection \ref{subsection:hela_dataset}). (a) Fluorescence micrograph with cytoplasm and nuclei stains pseudo-coloured red and blue respectively, $x_n$. (b) Annotations of $x_n$ corresponding to the cytoplasm instance mask  $y^2_n$ and (c) nuclei instance masks $y^1_n$. Scale bar 200 $\mu$m.}
    \label{fig:hela_sample}
\end{figure}

We have used this published dataset \citep{rodare_HeLaCytoNuc24} that consists of 2676 16-bit RGB fluorescence microscopy images of resolution 1040 x 1392, in which both the nuclei and cytoplasm channels were imaged. Here, the immortalised HeLa (ATCC-CCL-2) cell line commonly used in the laboratory was grown in cell culture, then fixed and stained with fluorescent dyes revealing nuclei and cytoplasm to be imaged using fluorescent microscopy.
We have chosen to work with images of optimal resolution (validated based on a quantifiable information loss using Dice-Coefficient \citep{Dice1945, Dice1948} between full-resolution and upscaled from a lower resolution images), at \(\frac{1}{2}\) of the original resolution along both the \emph{x} and \emph{y} axes, created from the raw 16-bit images. 
Corresponding to the images \{$X_n$\} are instance masks for nuclei \{$Y^1_n$\} and cytoplasm \{$Y^2_n$\}. The dataset split available for train, validation and test at 0.7:0.2:0.1 ratio was used. 
The schematic of the dataset consisting of the nuclei and cytoplasm masks is shown in Figure \ref{fig:hela_sample}.

For further dataset creation and acquisition details, please see \citep{rodare_HeLaCytoNuc24}.

\subsection{VACVPlaque Dataset}
\label{subsection:plaque_dataset}
\begin{figure}[!t]
    \centering
    \includegraphics[width=0.75\columnwidth]{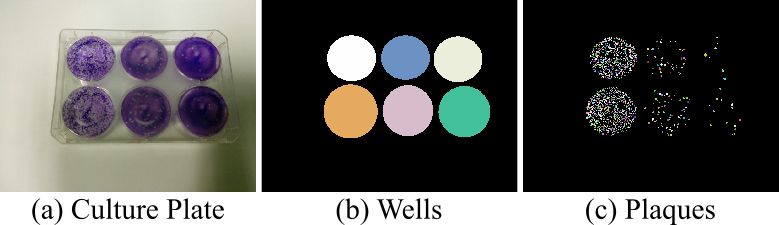}
    \caption{VACVPlaque \citep{rodare_VACVPlaque24} dataset (Subsection \ref{subsection:plaque_dataset}). (a) RGB mobile photographs of plaques within 6-well tissue culture plate, $x_n$. (b) Annotations of $x_n$ corresponding to the wells instance mask $y^2_n$ and (c) plaques instance mask $y^1_n$.}
    \label{fig:plaque_sample}
\end{figure}

The second published dataset used was \citep{rodare_VACVPlaque24} which contains digital photographs of 6-well tissue culture plates containing VACV \citep{Vaccinia09} virological plaque formations. The dataset consists of 211 8-bit RGB images of resolution 2448 x 3264 pixels (\emph{H} x \emph{W}). Corresponding to the images, \{$X_n$\} we also have instance masks for plaques \{$Y^1_n$\} and wells \{$Y^2_n$\}. The dataset split available for train, validation and test at 0.7:0.2:0.1 ratio was used. The schematic of the dataset consisting of the well and plaque masks is shown in Figure \ref{fig:plaque_sample}.
Here too we choose an optimal resolution (validated based on a quantifiable information loss using Dice-Coefficient \citep{Dice1945, Dice1948} between full-resolution and upscaled from a lower resolution images) to contain just enough resolution to detect the smallest plaques at \(\frac{1}{4}\) of the original resolution along both the \emph{x} and \emph{y} axes.

For further dataset creation and acquisition details, please see \citep{rodare_VACVPlaque24}.

\subsection{Experimental Setup}
\label{subsection:experimental_setup}
The skeleton of our code is the \href{https://github.com/stardist}{StarDist repository} \citep{Stardist18}. We follow the \href{https://cellpose.readthedocs.io/en/latest/}{Cellpose usage tutorials} \citep{Cellpose21} for implementation. We modified modules to incorporate changes related to the architecture, loss computations, dataset generation and predictions. We also implemented our work in Tensorflow \citep{Tensorflow15} for StarDist (marked SD) and PyTorch \citep{PyTorch19} for Cellpose (marked CP). 

For SD, based on our investigation about the ability of star-convex polygons to accurately mimic the outline of our objects, we have chosen the number of ($r^k_{i\,j}$), \emph{K} = 32 and $\lambda_1 = 0.2$, $\lambda_2 = 0.0001$, $\lambda_3 = 1$, $\epsilon = 1\mathrm{e}{-7}$ (Section \ref{sec:methods}) for all of our experiments. To optimise the model’s parameters, we employed the Adam optimiser \citep{Adam15} with an initial learning rate of 0.0003. Then, we decreased the learning rate according to ReduceLROnPlateau \citep{ReduceLROnPlateau16} with a decay rate of 0.5 with a patience of 40. The data for each epoch was random patches of size (256 x 256) from the training images with atmost 0.1 fraction of patches containing only background pixels. We used a batch size of 32, and the model’s parameters were updated with mini-batch gradient descent \citep{MiniBatch11}.
For Cellpose, all defaults training configurations including the default option of filtering out images containing less than 5 masks was used. The batch size was equal to 8, and the model’s parameters were updated with mini-batch gradient descent \citep{MiniBatch11}. The models were used with \href{https://www.python.org/downloads/release/python-3918/}{Python 3.9.18}. \\
In line with the task that the \href{https://cellpose.readthedocs.io/en/latest/models.html#full-built-in-models}{foundational models of Cellpose} was trained for, without finetuning, "cyto3" especially performed poorly on our tasks. We have considered versions in our comparison, that are trained fully from randomly initialised weights on our datasets or finetuned based on pretrained weights on our datasets. Also, since Cellpose is applicable for cellular objects, we refrain from using it with VACVPlaque (Subsection \ref{subsection:plaque_dataset}) which does not meet this criterion.\\ 
In our experiments, we have trained our baselines and branched architectures starting from randomly initialised weights for 400 epochs (marked (RI)). For baselines finetuned from pretrained weights, we have trained for 300 epochs (marked (FT)). In both cases, the epochs were sufficient for convergence. In our tables, \ref{tab:joint_tp_rate_hela_branched_foundational_performance}-\ref{tab:vacv_foundational_performance}, "\#Parameters" and "Training" for two-shot approaches SD and CP indicate a summation of parameter count and training time from two experiments set up for detecting the two categories of objects. All experiments were conducted on a NVIDIA RTX A6000 48GB GPU device. \\
All reproducible experimental code for all published values will be made available on GitHub under the \href{https://opensource.org/license/mit}{MIT} license as part of the camera-ready.

\section{Results}
\label{sec:results}
\begin{figure}[!t]
    \centering
    \includegraphics[width=0.75\columnwidth]{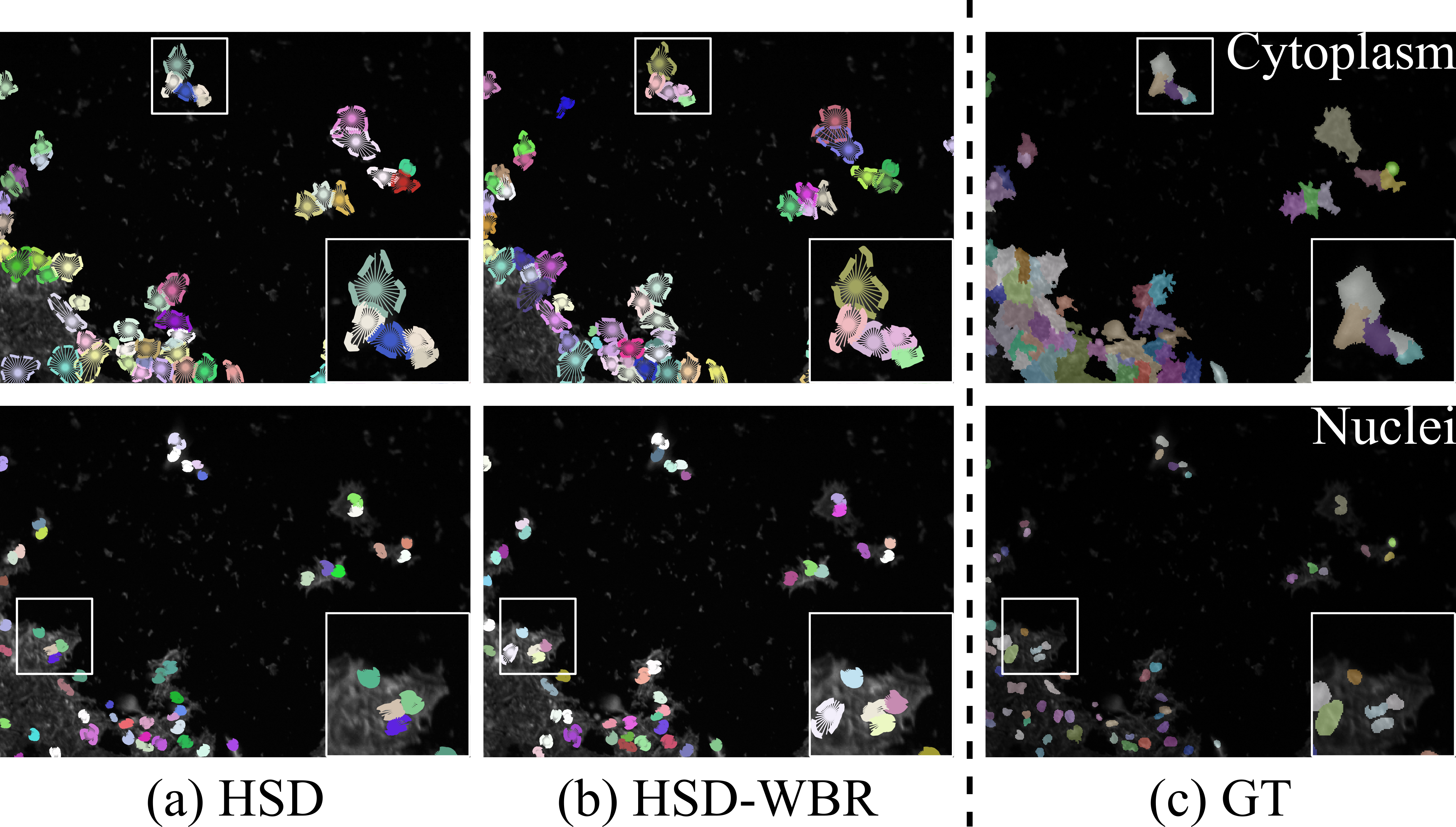}
    \caption{HeLaCytoNuc Single-shot instance segmentation results (Sections \ref{sec:experiments} and \ref{sec:discussion}), Predicted and Ground Truth (GT) (First row - Cytoplasm, Second row - Nuclei). Results for (a) HSD and (b) HSD-WBR (Section \ref{sec:methods}).}
    \label{fig:predictions_hela_branched}
\end{figure}

\begin{figure}[!t]
    \centering
    \includegraphics[width=0.75\columnwidth]{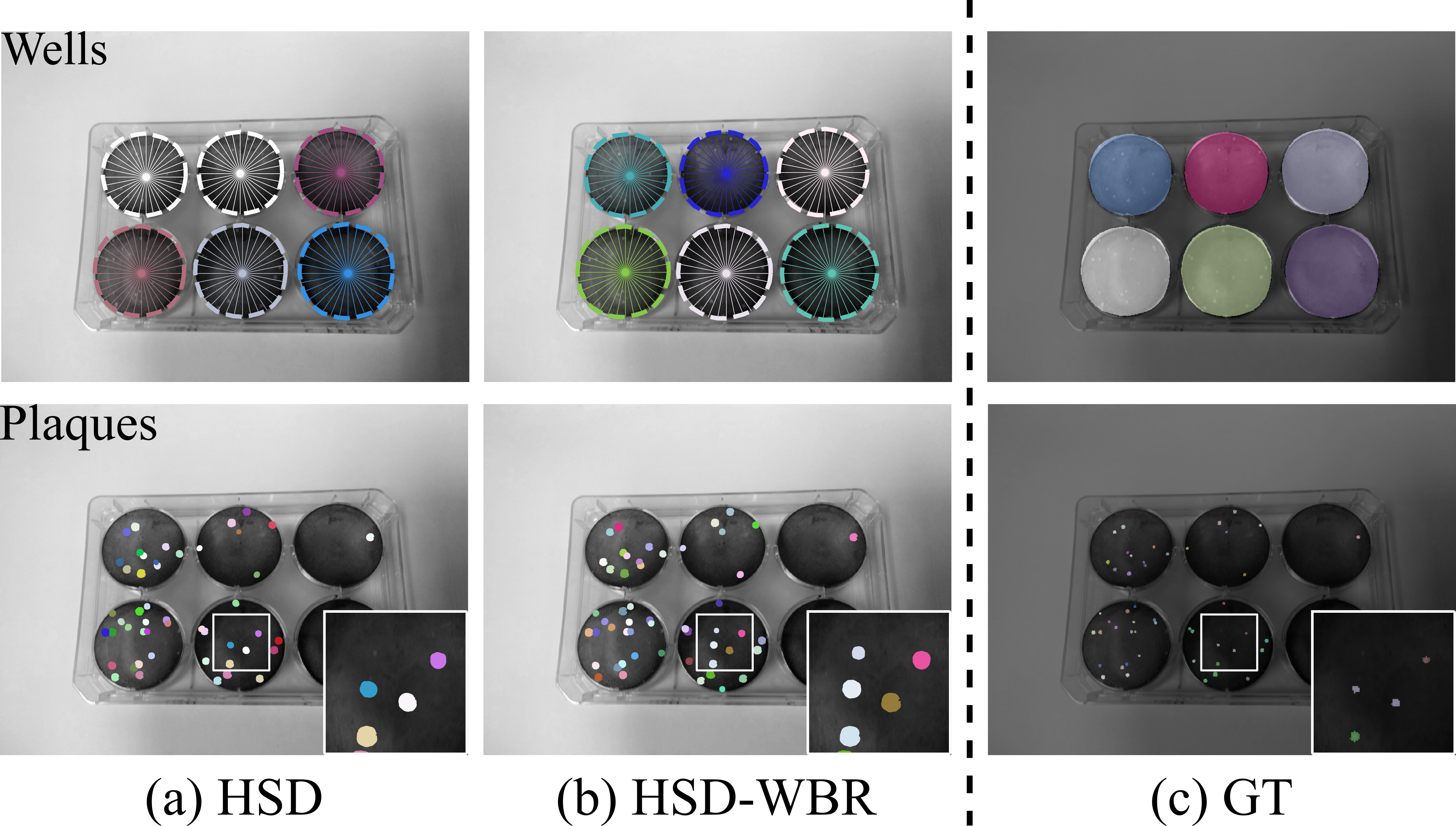}
    \caption{VACVPlaque Single-shot instance segmentation results (Sections \ref{sec:experiments} and \ref{sec:discussion}), Predicted and Ground Truth (GT) (First row - Wells, Second row - Plaques). Results for (a) HSD and (b) HSD-WBR (Section \ref{sec:methods}).}.  
    \label{fig:predictions_vacv_branched}
\end{figure}

Our findings suggest that on task-relevant evaluation criteria, $JTPR$, our methods HSD and HSD-WBR perform superior, joint best or a close second in Tables \ref{tab:joint_tp_rate_hela_branched_foundational_performance}, \ref{tab:joint_tp_rate_vacv_branched_foundational_performance}. 
From Figures \ref{fig:predictions_hela_branched}-\ref{fig:predictions_vacv_branched} and Tables \ref{tab:hela_branched_performance}-\ref{tab:vacv_branched_performance} we see that HSD-WBR improves TP at the cost of FP as compared to HSD which in case of biomedical objects is often preferable. This is resonated by the results from Tables \ref{tab:joint_tp_rate_hela_branched_foundational_performance}-\ref{tab:joint_tp_rate_vacv_branched_foundational_performance} suggesting the superior performance of HSD-WBR.

\begin{figure}[!t]
    \centering
    \includegraphics[width=0.75\columnwidth]{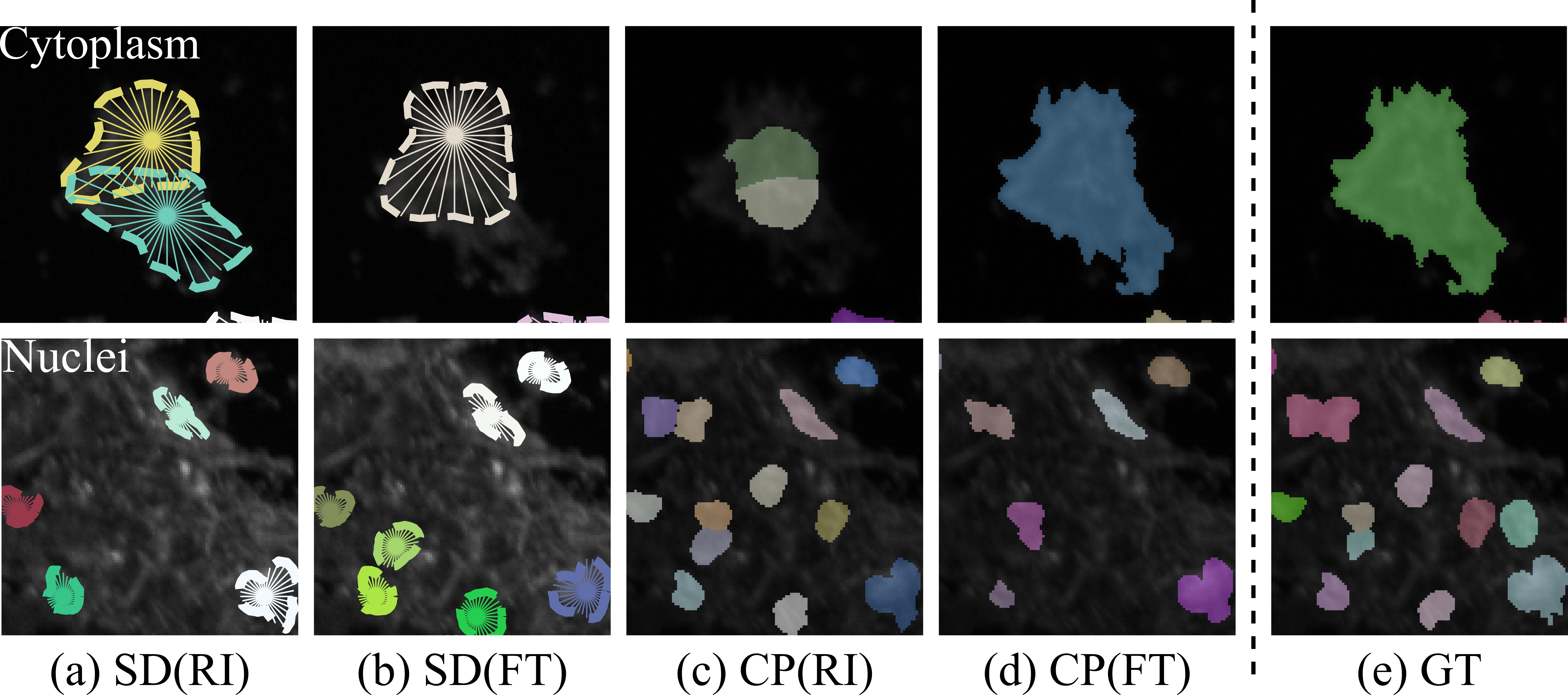}
    \caption{HeLaCytoNuc Two-shot instance segmentation results (Sections \ref{sec:experiments} and \ref{sec:discussion}), Predicted and Ground Truth (GT) (First row - Cytoplasm, Second row - Nuclei). Results for SOTA architectures (a), (b) SD and (c), (d) CP trained from randomly initialised weights \emph{(RI)} and finetuned starting from pretrained weights \emph{(FT)} (Section \ref{sec:experiments}).}
    \label{fig:predictions_hela_foundational}
\end{figure}

\begin{figure}[!t]
    \centering
    \includegraphics[width=0.75\columnwidth]{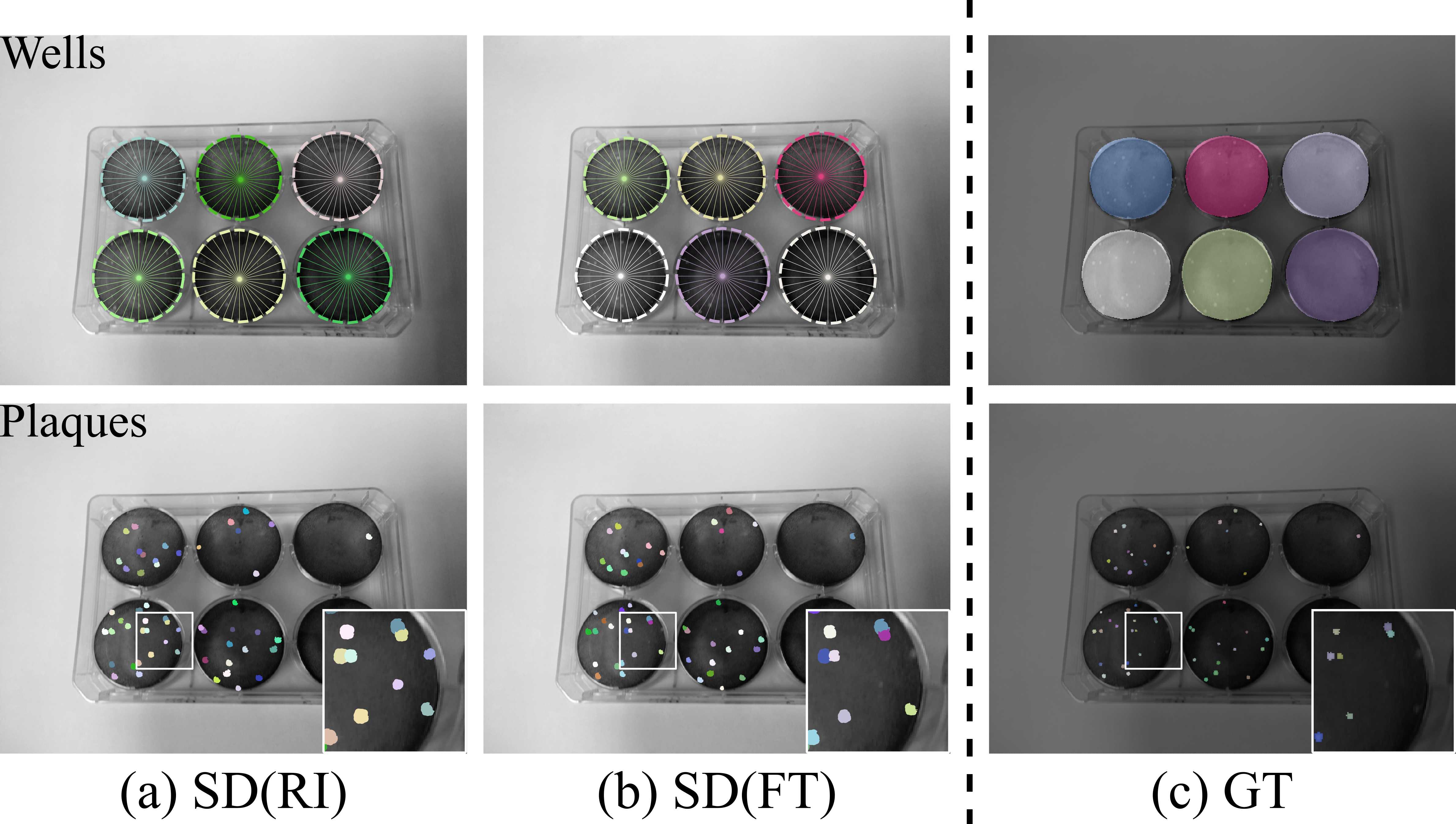}
    \caption{VACVPlaque Two-shot instance segmentation results (Sections \ref{sec:related_works} and \ref{sec:experiments}), Predicted and Ground Truth (GT) (First row - Wells, Second row - Plaques). Results for SOTA architecture SD trained from (a) randomly initialised weights \emph{(RI)} and (b) finetuned starting from pretrained weights \emph{(FT)} (Section \ref{sec:experiments}).}
    \label{fig:predictions_vacv_foundational}
\end{figure}

Our experiments with $JTPR$ (Equations \ref{eq:jtp_rate_inner}-\ref{eq:jtp_rate_outer_vacv}), suggest $AP$ (Equation \ref{eq:ap}) is insufficient for capturing the performance on the spatially correlated instance segmentation. When seen in tandem with visual prediction results, from Figure \ref{fig:predictions_hela_foundational}, we see that CP (FT) seems to have performed the best while Table \ref{tab:hela_foundational_performance} ($AP$) tells a different story. Similarly, even though in Figure \ref{fig:predictions_vacv_foundational}, SD (FT) is visually the best, from Table \ref{tab:vacv_foundational_performance} we see that SD (RI) scores better. Furthermore, results on $IoU_R$ (Tables \ref{tab:joint_tp_rate_hela_branched_foundational_performance}-\ref{tab:hela_foundational_performance}) suggest that the performance of CP (FT) is much closer the the best performance and in line with the visual results. But even though $IoU_R$ captures the good performance of CP (FT), it is not very discriminative in performance quantification as seen from visual predictions. Only in Table \ref{tab:hela_foundational_performance}, for $IoU_R$, SD (FT) scores significantly above any other two-shot or single-shot architecture. In all other tables, even based on $IoU_R$, the results remain inconclusive. \\ 
Conversely, using $JTPR$ (Equations \ref{eq:jtp_rate_inner}-\ref{eq:jtp_rate_outer_vacv}), CP (FT) from Figure \ref{fig:predictions_hela_foundational} demonstrates a score much closer to the best performance in Table \ref{tab:joint_tp_rate_hela_branched_foundational_performance}.\\
For complete values of $JTPR$, $IoU_R$ and $AP$ at all $\tau$s and ablations results (based on these criteria) of $\lambda_1$, $\lambda_2$ and $\lambda_3$ on SD, HSD and HSD-WBR please see the Appendix (Section \ref{sec:appendix}). The experimental results in the Appendix have been reported on both datasets used here.

\begin{table}[h!]
  \caption{HeLaCytoNuc Single-shot existing metrics. Single-shot \emph{$IoU_R$} (Equation \ref{eq:iou-recall}) and \emph{$AP$} (Equation \ref{eq:ap}) reported at optimal ($\tau$), parameter count and training time on HeLaCytoNuc (Subsection \ref{subsection:hela_dataset}). Training here suggests that for single-shot, for each category of objects approximately half of the total training time is needed. \textbf{Best}. }
  \centering
  \resizebox{0.75\linewidth}{!}{
  \begin{tabular}{cccccc}
   \toprule
    \multirow{2}{*}{Task} & \multirow{2}{*}{Architecture} & \multirow{2}{*}{$IoU_R$ $\uparrow$} & \multirow{2}{*}{$AP$ $\uparrow$} & \#Parameters $\downarrow$ & Training $\downarrow$ \\
    & & & & ($\approx$ millions) & (mins) \\
    \midrule
    \multirow{2}{*}{\textbf{Nuclei}} & \emph{HSD-WBR} (ours) & \textbf{0.698} & \textbf{0.866} & $\mathbf{30.2}$ & 207\slash 2 \\
    & \emph{HSD} (ours) & 0.694 & 0.857 & $\mathbf{30.2}$ & \textbf{206\slash 2} \\
    \midrule
    \multirow{2}{*}{\textbf{Cytoplasm}} & \emph{HSD-WBR} (ours) & \textbf{0.567} & \textbf{0.725} & $\mathbf{30.2}$ & 207\slash 2 \\
    & \emph{HSD} (ours) & 0.549 & \textbf{0.725} & $\mathbf{30.2}$ & \textbf{206\slash 2} \\
    \bottomrule
    \end{tabular}
    }
    \label{tab:hela_branched_performance}
\end{table}

\begin{table}[h!]
   \caption{VACVPlaque Single-shot existing metrics. Single-shot \emph{$IoU_R$} (Equation \ref{eq:iou-recall}) and \emph{$AP$} (Equation \ref{eq:ap}) reported at optimal ($\tau$), parameter count and training time on VACVPlaque (Subsection \ref{subsection:plaque_dataset}). Training here suggests that for single-shot, for each category of objects approximately half of the total training time is needed. \textbf{Best}. }
  \centering
  \resizebox{0.75\linewidth}{!}{
  \begin{tabular}{cccccc}
   \toprule
    \multirow{2}{*}{Task} & \multirow{2}{*}{Architecture} & \multirow{2}{*}{$IoU_R$ $\uparrow$} & \multirow{2}{*}{$AP$ $\uparrow$} & \#Parameters $\downarrow$ & Training $\downarrow$ \\
    & & & & ($\approx$ millions) & (mins) \\
    \midrule
    \multirow{2}{*}{\textbf{Plaques}} & \emph{HSD-WBR} (ours) & \textbf{0.436} & \textbf{0.645} & $\mathbf{30.2}$ & \textbf{17\slash 2} \\
    & \emph{HSD} (ours) & 0.420 & 0.618 & $\mathbf{30.2}$ & \textbf{17\slash 2} \\
    \midrule
    \multirow{2}{*}{\textbf{Wells}} & \emph{HSD-WBR} (ours) & \textbf{0.957} & 0.920 & $\mathbf{30.2}$ & \textbf{17\slash 2} \\
    & \emph{HSD} (ours) & 0.954 & \textbf{0.967} & $\mathbf{30.2}$ & \textbf{17\slash 2} \\
  \bottomrule
  \end{tabular}
  }
  \label{tab:vacv_branched_performance}
\end{table}

\begin{table}[h!]
  \caption{HeLaCytoNuc Two-shot existing metrics. Two-shot \emph{$IoU_R$} (Equation \ref{eq:iou-recall}) and \emph{$AP$} (Equation \ref{eq:ap}) reported at optimal ($\tau$) on HeLaCytoNuc (Subsection \ref{subsection:hela_dataset}). \textbf{Best}. \underline{Second best}.}
  \centering
  \resizebox{0.75\linewidth}{!}{
  \begin{tabular}{cccccc}
    \toprule
    \multirow{2}{*}{Task} & \multirow{2}{*}{Architecture} & \multirow{2}{*}{$IoU_R$ $\uparrow$} & \multirow{2}{*}{$AP$ $\uparrow$} & \#Parameters $\downarrow$ & Training $\downarrow$ \\
    & & & & ($\approx$ millions) & (mins) \\
    \midrule
    \multirow{4}{*}{\textbf{Nuclei}} & \emph{SD} (RI)\citep{Stardist18} & \underline{0.696} & \underline{0.858} & \underline{22.2} & \underline{131} \\
    & \emph{SD} (FT)\citep{Stardist18} & \textbf{0.814} & \textbf{0.860} & \underline{22.2} & 147 \\
    & \emph{Cellpose} (RI)\citep{Cellpose21} & 0.686 & 0.837 & \textbf{6.6} & 135 \\
    & \emph{Cellpose} (FT)\citep{Cellpose21} & 0.621 & 0.759 & \textbf{6.6} & \textbf{103} \\
    \midrule
    \multirow{4}{*}{\textbf{Cytoplasm}} & \emph{SD} (RI)\citep{Stardist18} & 0.511 & \underline{0.697} & \underline{22.2} & 146 \\
    & \emph{SD} (FT)\citep{Stardist18} & \textbf{0.579} & \textbf{0.769} & \underline{22.2} & \textbf{84} \\ 
    & \emph{Cellpose} (RI)\citep{Cellpose21} & 0.534 & 0.692 & \textbf{6.6} & 145 \\
    & \emph{Cellpose} (FT)\citep{Cellpose21} & \underline{0.538} & 0.696 & \textbf{6.6} & \underline{137} \\
  \bottomrule
  \end{tabular}
  }
  \label{tab:hela_foundational_performance}
\end{table}

\begin{table}[h!]
  \caption{VACVPlaque Two-shot existing metrics. Two-shot \emph{$IoU_R$} (Equation \ref{eq:iou-recall}) and \emph{$AP$} (Equation \ref{eq:ap}) reported at optimal ($\tau$) on
  VACVPlaque (Subsection \ref{subsection:plaque_dataset}) for SD \citep{Stardist18}. Experiments with Cellpose were omitted due to to its applicability only to cellular objects. \textbf{Best}.}
  \centering
  \resizebox{0.75\linewidth}{!}{
  \begin{tabular}{cccccc}
   \toprule
    \multirow{2}{*}{Task} & \multirow{2}{*}{Architecture} & \multirow{2}{*}{$IoU_R$ $\uparrow$} & \multirow{2}{*}{$AP$ $\uparrow$} & \#Parameters $\downarrow$ & Training $\downarrow$ \\
    & & & & ($\approx$ millions) & (mins) \\
    \midrule
    \multirow{2}{*}{\textbf{Plaques}} & \emph{SD} (RI)\citep{Stardist18} & \textbf{0.494} & \textbf{0.730} & \textbf{22.2} & 13 \\
    & \emph{SD} (FT)\citep{Stardist18} & 0.357 & 0.566 & \textbf{22.2} & \textbf{8} \\
    \midrule
    \multirow{2}{*}{\textbf{Wells}} & \emph{SD} (RI)\citep{Stardist18} & \textbf{0.955} & \textbf{1.000} & \textbf{22.2} & 13 \\
    & \emph{SD} (FT)\citep{Stardist18} & 0.947 & 0.992 & \textbf{22.2} & \textbf{8} \\
  \bottomrule
  \end{tabular}
  }
  \label{tab:vacv_foundational_performance}
\end{table}

\section{Discussion}
\label{sec:discussion}
Objects in biomedical images are often spatially correlated. Detecting them in a single-shot manner is often both desirable and can exploit the correlation as an important prior. Furthermore, employing architectures for single-shot detection can improve computational efficiency, which bears important consequences for the sustainability and economic viability of DL solutions \citep{Liu24, Thompson20}. Here we show, that for the tasks at hand, with few exceptions the branched single-shot architectures were more efficient without compromising performance (Tables \ref{tab:joint_tp_rate_hela_branched_foundational_performance}-\ref{tab:joint_tp_rate_vacv_branched_foundational_performance}).\\
These exceptions include $SD(FT)$ training times (Tables \ref{tab:joint_tp_rate_vacv_branched_foundational_performance}, \ref{tab:hela_foundational_performance}-\ref{tab:vacv_foundational_performance}). They likely stem from selective parameter fine-tuning of the highly optimised SD architecture. Furthermore, when evaluated on the task-relevant $JTPR$ criteria (Tables \ref{tab:joint_tp_rate_hela_branched_foundational_performance}-\ref{tab:joint_tp_rate_vacv_branched_foundational_performance}) the single shot models we propose outperformed the SOTA (Tables \ref{tab:joint_tp_rate_hela_branched_foundational_performance}-\ref{tab:joint_tp_rate_vacv_branched_foundational_performance}) with \textit{highest} JTPR, \textit{second-lowest} \#Parameters and \textit{lowest} training times. It is tempting to speculate that the spatial correlation prior allows HSD and HSD-WBR to outperform even specialist models like CP in this task. If true, this would suggest that a more precise end-to-end formulation of tasks is preferable to a specialist model if the evaluation criteria are in line with the end goal. We aim to investigate this in future.

\section{Acknowledgements \& Funding Information}
This work was partially funded by the Center for Advanced Systems Understanding (CASUS) which is financed by Germany’s Federal Ministry of Education and Research (BMBF) and by the Saxon Ministry for Science, Culture, and Tourism (SMWK) with tax funds on the basis of the budget approved by the Saxon State Parliament. The authors acknowledge the financial support by the Federal Ministry of Education and Research of Germany and by Sächsische Staatsministerium für Wissenschaft, Kultur und Tourismus in the programme Center of Excellence for AI-research „Center for Scalable Data Analytics and Artificial Intelligence Dresden/Leipzig“, project identification number: ScaDS.AI. We acknowledge the kind support of Prof. Dr. Jason Mercer, Dr. Moona Huttunen and the former InfectX consortium for access to the raw assay plates and micrographs.\\

TD and AY thank AU for his support in setting up the compute resource used for the experiments in this work and also for the preparation of the experiment-ready data for the HeLaCytoNuc dataset. The authors thank Maria Wyrzykowska, former Masters Student at Helmholtz-Zentrum Dresden-Rossendorf e. V. (HZDR) and University of Wrocław for her initial experiments with Cellpose as part of her Masters thesis. The authors thank Subasini Thangamani, former Masters Student at HZDR and Technical University of Dresden for her initial experiments using HSD and for preparation of the VACVPlaque dataset as part of her Masters Thesis.

\bibliographystyle{unsrtnat}
\bibliography{references}

\section{Appendix}
\label{sec:appendix}
The aim of the appendix is to provide readers with a deeper insight into our experimental results.
\subsection{Detailed experimental results for all $\tau$s}
\label{subsection:detailed_results_all_taus}
From Tables \ref{tab:sup_iou_hela_branched_performance}-\ref{tab:sup_ap_vacv_foundational_performance}, readers are welcome to choose to arrive at different conclusions about the optimal architecture and training to use. This can be done by considering different cross-sections of the tables, focusing on an a specific \textit{$\tau$} or any other averaging method in conjunction with the required compute and training times mentioned in the tables in the main script. \\
In Tables \ref{tab:sup_iou_hela_branched_performance}-\ref{tab:sup_ap_vacv_foundational_performance}, all odd-numbered tables relate to experiments with HeLaCytoNuc and all even-numbered tables relate to VACVPlaque.

\begin{table}[h!]
  \caption{HeLaCytoNuc single-shot \emph{$IoU_R$} (Equation \ref{eq:iou-recall}) at different thresholds ($\tau$)}
  \label{tab:sup_iou_hela_branched_performance}
  \centering
  \resizebox{\linewidth}{!}{
  \begin{tabular}{ccccccccccc}
   \toprule
    & &  \multicolumn{9}{c}{$\tau$} \\
    \cmidrule(r){3-11}
    Task & Architecture & 0.1 & 0.2 & 0.3 & 0.4 & 0.5 & 0.6 & 0.7 & 0.8 & 0.9 \\
    \midrule
    \multirow{2}{*}{\textbf{Nuclei}} & \emph{HSD-WBR}(ours) & \textbf{0.698} & \textbf{0.698} & \textbf{0.697} & \textbf{0.693} & \textbf{0.680} & \textbf{0.654} & 0.599 & 0.461 & 0.108 \\
    & \emph{HSD}(ours) & 0.694 & 0.694 & 0.693 & 0.689 & 0.677 & \textbf{0.654} & \textbf{0.601} & \textbf{0.468} & \textbf{0.117} \\
    \midrule
    \multirow{2}{*}{\textbf{Cytoplasm}} & \emph{HSD-WBR}(ours) & \textbf{0.567} & \textbf{0.566} & \textbf{0.562} & \textbf{0.549} & \textbf{0.518} & \textbf{0.465} & \textbf{0.376} & \textbf{0.232} & \textbf{0.039} \\
    & \emph{HSD}(ours) & 0.549 & 0.548 & 0.544 & 0.532 & 0.502 & 0.450 & 0.363 & 0.217 & 0.030 \\
    \bottomrule
\end{tabular}
}
\end{table}

\begin{table}[h!]
  \caption{VACVPlaque single-shot \emph{$IoU_R$} (Equation \ref{eq:iou-recall}) at different thresholds ($\tau$)}
  \label{tab:sup_iou_vacv_branched_performance}
  \centering
  \resizebox{\linewidth}{!}{
  \begin{tabular}{ccccccccccc}
   \toprule
    & &  \multicolumn{9}{c}{$\tau$} \\
    \cmidrule(r){3-11}
    Task & Architecture & 0.1 & 0.2 & 0.3 & 0.4 & 0.5 & 0.6 & 0.7 & 0.8 & 0.9 \\
    \midrule
    \multirow{2}{*}{\textbf{Plaques}} & \emph{HSD-WBR}(ours) & \textbf{0.436} & \textbf{0.434} & \textbf{0.427} & \textbf{0.409} & \textbf{0.376} & \textbf{0.298} & \textbf{0.186} & \textbf{0.076} & \textbf{0.013} \\
    & \emph{HSD}(ours) & 0.420 & 0.418 & 0.410 & 0.392 & 0.358 & 0.282 & 0.170 & 0.066 & 0.009 \\
    \midrule
    \multirow{2}{*}{\textbf{Wells}} & \emph{HSD-WBR}(ours) & \textbf{0.957} & \textbf{0.957} & \textbf{0.957} & \textbf{0.957} & \textbf{0.957} & \textbf{0.957} & \textbf{0.957} & \textbf{0.953} & \textbf{0.951} \\
    & \emph{HSD}(ours) & 0.954 & 0.954 & 0.954 & 0.954 & 0.954 & 0.954 & 0.954 & 0.948 & 0.948 \\
  \bottomrule
  \end{tabular}
  }
\end{table}

\begin{table}[h!]
  \caption{HeLaCytoNuc two-shot \emph{$IoU_R$} (Equation \ref{eq:iou-recall}) at different thresholds ($\tau$)}
  \label{tab:sup_iou_hela_foundational_performance}
  \centering
  \resizebox{\linewidth}{!}{
  \begin{tabular}{ccccccccccc}
    \toprule
    & &  \multicolumn{9}{c}{$\tau$} \\
    \cmidrule(r){3-11}
    Task & Architecture & 0.1 & 0.2 & 0.3 & 0.4 & 0.5 & 0.6 & 0.7 & 0.8 & 0.9 \\
    \midrule
    \multirow{4}{*}{\textbf{Nuclei}} & \emph{SD} (RI) \citep{Stardist18} & \underline{0.696} & \underline{0.696} & \underline{0.695} & \underline{0.691} & \underline{0.679} & \underline{0.658} & \underline{0.610} & 0.488 & 0.121 \\
    & \emph{SD} (FT) \citep{Stardist18} & \textbf{0.814} & \textbf{0.814} & \textbf{0.814} & \textbf{0.812} & \textbf{0.808} & \textbf{0.798} & \textbf{0.779} & \textbf{0.729} & \textbf{0.494} \\
    & \emph{Cellpose} (RI) \citep{Cellpose21} & 0.686 & 0.686 & 0.685 & 0.682 & 0.672 & 0.652 & 0.609 & \underline{0.495} & \underline{0.161} \\
    & \emph{Cellpose} (FT) \citep{Cellpose21} & 0.621 & 0.621 & 0.621 & 0.618 & 0.611 & 0.594 & 0.555 & 0.444 & 0.140 \\
    \midrule
    \multirow{4}{*}{\textbf{Cytoplasm}} & \emph{SD} (RI) \citep{Stardist18} & 0.511 & 0.510 & 0.507 & 0.497 & 0.475 & 0.431 & \underline{0.353} & 0.217 & \underline{0.030} \\
    & \emph{SD} (FT) \citep{Stardist18} & \textbf{0.579} & \textbf{0.578} & \textbf{0.574} & \textbf{0.559} & \textbf{0.520} & 0.451 & 0.336 & 0.165 & 0.009 \\ 
    & \emph{Cellpose} (RI) \citep{Cellpose21} & 0.534 & 0.534 & 0.532 & 0.524 & 0.505 & \underline{0.469} & \textbf{0.414} & \underline{0.316} & \textbf{0.144} \\
    & \emph{Cellpose} (FT) \citep{Cellpose21} & \underline{0.538} & \underline{0.537} & \underline{0.535} & \underline{0.527} & \underline{0.508} & \textbf{0.471} & \textbf{0.414} & \textbf{0.317} & \textbf{0.144} \\
  \bottomrule
  \end{tabular}
  }
\end{table}

\begin{table}[h!]
  \caption{VACVPlaque two-shot \emph{$IoU_R$} (Equation \ref{eq:iou-recall}) at different thresholds ($\tau$)}
  \label{tab:sup_iou_vacv_foundational_performance}
  \centering
  \resizebox{\linewidth}{!}{
  \begin{tabular}{ccccccccccc}
   \toprule
    & &  \multicolumn{9}{c}{$\tau$} \\
    \cmidrule(r){3-11}
    Task & Architecture & 0.1 & 0.2 & 0.3 & 0.4 & 0.5 & 0.6 & 0.7 & 0.8 & 0.9 \\
    \midrule
    \multirow{2}{*}{\textbf{Plaques}} & \emph{SD} (RI) \citep{Stardist18} & \textbf{0.494} & \textbf{0.493} & \textbf{0.488} & \textbf{0.474} & \textbf{0.444} & \textbf{0.366} & \textbf{0.243} & \textbf{0.106} & \textbf{0.019} \\
    & \emph{SD} (FT) \citep{Stardist18} & 0.357 & 0.356 & 0.350 & 0.335 & 0.307 & 0.242 & 0.147 & 0.056 & 0.008 \\
    \midrule
    \multirow{2}{*}{\textbf{Wells}} & \emph{SD} (RI) \citep{Stardist18} & \textbf{0.955} & \textbf{0.955} & \textbf{0.955} & \textbf{0.955} & \textbf{0.955} & \textbf{0.955} & \textbf{0.955} & \textbf{0.955} & \textbf{0.944} \\
    & \emph{SD} (FT) \citep{Stardist18} & 0.947 & 0.947 & 0.954 & 0.947 & 0.947 & 0.947 & 0.947 & 0.947 & 0.926 \\
  \bottomrule
  \multicolumn{11}{p{350pt}}{Experiments with Cellpose were omitted due to to its applicability only to cellular objects.}
  \end{tabular}
  }
\end{table}

\begin{table}[h!]
  \caption{HeLaCytoNuc single-shot \emph{AP} (Equation \ref{eq:ap}) at different thresholds ($\tau$)}
  \label{tab:sup_ap_hela_branched_performance}
  \centering
  \resizebox{\linewidth}{!}{
  \begin{tabular}{ccccccccccc}
    \toprule
    & &  \multicolumn{9}{c}{$\tau$} \\
    \cmidrule(r){3-11}
    Task & Architecture & 0.1 & 0.2 & 0.3 & 0.4 & 0.5 & 0.6 & 0.7 & 0.8 & 0.9 \\
    \midrule
    \multirow{2}{*}{\textbf{Nuclei}} & \emph{HSD-WBR}(ours) & \textbf{0.866} & \textbf{0.864} & \textbf{0.855} & \textbf{0.836} & \textbf{0.787} & 0.713 & 0.592 & 0.384 & 0.066 \\
    & \emph{HSD}(ours) & 0.857 & 0.855 & 0.847 & 0.828 & 0.783 & \textbf{0.715} & \textbf{0.598} & \textbf{0.393} & \textbf{0.072} \\
    \midrule
    \multirow{2}{*}{\textbf{Cytoplasm}} & \emph{HSD-WBR}(ours) & \textbf{0.725} & 0.716 & 0.694 & 0.640 & 0.553 & 0.442 & \textbf{0.311} & \textbf{0.161} & \textbf{0.022} \\
    & \emph{HSD}(ours) & \textbf{0.725} & \textbf{0.718} & \textbf{0.697} & \textbf{0.645} & \textbf{0.556} & \textbf{0.444} & 0.310 & 0.155 & 0.018 \\    
  \bottomrule
  \end{tabular}
  }
\end{table}

\begin{table}[h!]
  \caption{VACVPlaque single-shot \emph{AP} (Equation \ref{eq:ap}) at different thresholds ($\tau$)}
  \label{tab:sup_ap_vacv_branched_performance}
  \centering
  \resizebox{\linewidth}{!}{
  \begin{tabular}{ccccccccccc}
    \toprule
    & &  \multicolumn{9}{c}{$\tau$} \\
    \cmidrule(r){3-11}
    Task & Architecture & 0.1 & 0.2 & 0.3 & 0.4 & 0.5 & 0.6 & 0.7 & 0.8 & 0.9 \\
    \midrule
    \multirow{2}{*}{\textbf{Plaques}} & \emph{HSD-WBR}(ours) & \textbf{0.645} & \textbf{0.631} & \textbf{0.588} & \textbf{0.521} & \textbf{0.434} & \textbf{0.288} & \textbf{0.148} & \textbf{0.051} & \textbf{0.007} \\
    & \emph{HSD}(ours) & 0.618 & 0.604 & 0.561 & 0.495 & 0.407 & 0.271 & 0.135 & 0.045 & 0.005 \\
    \midrule
    \multirow{2}{*}{\textbf{Wells}} & \emph{HSD-WBR}(ours) & 0.920 & 0.920 & 0.920 & 0.920 & 0.920 & 0.920 & 0.920 & 0.911 & 0.906  \\
    & \emph{HSD}(ours) & \textbf{0.967} & \textbf{0.967} & \textbf{0.967} & \textbf{0.967} & \textbf{0.967} & \textbf{0.967} & \textbf{0.967} & \textbf{0.952} & \textbf{0.952} \\
  \bottomrule
  \end{tabular}
  }
\end{table}

\begin{table}[h!]
  \caption{HeLaCytoNuc two-shot \emph{AP} (Equation \ref{eq:ap}) at different thresholds ($\tau$)}
  \label{tab:sup_ap_hela_foundational_performance}
  \centering
  \resizebox{\linewidth}{!}{
  \begin{tabular}{ccccccccccc}
    \toprule
    & &  \multicolumn{9}{c}{$\tau$} \\
    \cmidrule(r){3-11}
    Task & Architecture & 0.1 & 0.2 & 0.3 & 0.4 & 0.5 & 0.6 & 0.7 & 0.8 & 0.9 \\
    \midrule
    \multirow{4}{*}{\textbf{Nuclei}} & \emph{SD} (RI) \citep{Stardist18} & \underline{0.858} & \underline{0.855} & \underline{0.848} & \underline{0.829} & \underline{0.785} & \underline{0.722} & 0.614 & 0.420 & 0.074 \\
    & \emph{SD} (FT) \citep{Stardist18} & \textbf{0.860} & \textbf{0.860} & \textbf{0.858} & \textbf{0.851} & \textbf{0.835} & \textbf{0.804} & \textbf{0.759} & \textbf{0.663} & \textbf{0.357} \\
    & \emph{Cellpose} (RI) \citep{Cellpose21} & 0.837 & 0.836 & 0.831 & 0.816 & 0.776 & 0.716 & \underline{0.618} & \underline{0.432} & \underline{0.103} \\
    & \emph{Cellpose} (FT) \citep{Cellpose21} & 0.759 & 0.759 & 0.756 & 0.746 & 0.717 & 0.667 & 0.579 & 0.398 & 0.093 \\
    \midrule
    \multirow{4}{*}{\textbf{Cytoplasm}} & \emph{SD} (RI) \citep{Stardist18} & \underline{0.697} & \underline{0.691} & 0.673 & 0.630 & 0.559 & 0.457 & 0.326 & \underline{0.168} & 0.019 \\
    & \emph{SD} (FT) \citep{Stardist18} & \textbf{0.769} & \textbf{0.763} & \textbf{0.739} & \textbf{0.677} & 0.565 & 0.426 & 0.267 & 0.108 & 0.005 \\ 
    & \emph{Cellpose} (RI) \citep{Cellpose21} & 0.692 & 0.687 & 0.676 & 0.641 & \underline{0.578} & \underline{0.490} & \textbf{0.389} & \textbf{0.258} & \textbf{0.098} \\
    & \emph{Cellpose} (FT) \citep{Cellpose21} & 0.696 & \underline{0.691} & \underline{0.680} & \underline{0.644} & \textbf{0.582} & \textbf{0.491} & \underline{0.387} & \textbf{0.258} & \textbf{0.098} \\
  \bottomrule
  \end{tabular}
  }
\end{table}

\begin{table}[h!]
  \caption{VACVPlaque two-shot \emph{AP} (Equation \ref{eq:ap}) at different thresholds ($\tau$)}
  \label{tab:sup_ap_vacv_foundational_performance}
  \centering
  \resizebox{\linewidth}{!}{
  \begin{tabular}{ccccccccccc}
   \toprule
    & &  \multicolumn{9}{c}{$\tau$} \\
    \cmidrule(r){3-11}
    Task & Architecture & 0.1 & 0.2 & 0.3 & 0.4 & 0.5 & 0.6 & 0.7 & 0.8 & 0.9 \\
    \midrule
    \multirow{2}{*}{\textbf{Plaques}} & \emph{SD} (RI) \citep{Stardist18} & \textbf{0.730} & \textbf{0.721} & \textbf{0.690} & \textbf{0.630} & \textbf{0.538} & \textbf{0.373} & \textbf{0.202} & \textbf{0.073} & \textbf{0.011} \\
    & \emph{SD} (FT) \citep{Stardist18} & 0.566 & 0.555 & 0.523 & 0.463 & 0.383 & 0.257 & 0.130 & 0.042 & 0.005 \\
    \midrule
    \multirow{2}{*}{\textbf{Wells}} & \emph{SD} (RI) \citep{Stardist18} & \textbf{1.000} & \textbf{1.000} & \textbf{1.000} & \textbf{1.000} & \textbf{1.000} & \textbf{1.000} & \textbf{1.000} & \textbf{1.000} & \textbf{0.974} \\
    & \emph{SD} (FT) \citep{Stardist18} & 0.992 & 0.992 & 0.992 & 0.992 & 0.992 & 0.992 & 0.992 & 0.992 & 0.946 \\
  \bottomrule
  \multicolumn{11}{p{350pt}}{Experiments with Cellpose were omitted due to to its applicability only to cellular objects.}
  \end{tabular}
  }
\end{table}

\subsection{Ablation Study of $\lambda_1$, $\lambda_2$, $\lambda_3$}
\label{subsection:ablation_lambda}
In this section, Tables \ref{tab:hela_ablation_lambda1_iou_performance}-\ref{tab:hela_ablation_lambda3_joint_tp_rate_performance} relate to ablation experiments with the HeLaCytoNuc dataset and \ref{tab:vacv_ablation_lambda1_iou_performance}-\ref{tab:vacv_ablation_lambda3_joint_tp_rate_performance} relate to ablation experiments with the VACVPlaque dataset.\\

From Tables \ref{tab:hela_ablation_lambda1_iou_performance}-\ref{tab:hela_ablation_lambda2_ap_performance} and \ref{tab:hela_ablation_lambda1_joint_tp_rate_performance}-\ref{tab:hela_ablation_lambda2_joint_tp_rate_performance}, readers are welcome to deduce for each SD-based model, an appropriate combination of $\lambda_1$ and $\lambda_2$ to use for final experiments using HeLaCytoNuc. From Tables \ref{tab:hela_ablation_lambda3_iou_performance}-\ref{tab:hela_ablation_lambda3_ap_performance}, and \ref{tab:hela_ablation_lambda3_joint_tp_rate_performance} readers can deduce which of the $\lambda_3$ value to use for final experiments for HeLaCytoNuc. Naturally only the HSD-WBR is applicable here.\\
Similarly, from Tables \ref{tab:vacv_ablation_lambda1_iou_performance}-\ref{tab:vacv_ablation_lambda2_ap_performance} and \ref{tab:vacv_ablation_lambda1_joint_tp_rate_performance}-\ref{tab:vacv_ablation_lambda2_joint_tp_rate_performance} readers can deduce an appropriate combination of $\lambda_1$ and $\lambda_2$ to use for final experiments using VACVPlaque. And from Tables \ref{tab:vacv_ablation_lambda3_iou_performance}-\ref{tab:vacv_ablation_lambda3_ap_performance}, and \ref{tab:vacv_ablation_lambda3_joint_tp_rate_performance} an appropriate $\lambda_3$ value to use for final experiments for VACVPlaque. Also here as before, only the HSD-WBR is applicable.\\

In each of the experiments, exactly one $\lambda$ has been varied keeping everything else about the experiment constant.

\begin{table}[h!]
  \caption{HeLaCytoNuc \emph{$IoU_R$} (Equation \ref{eq:iou-recall}) at different thresholds ($\tau$) and $\lambda_1$.}
  \label{tab:hela_ablation_lambda1_iou_performance}
  \centering
  \resizebox{\linewidth}{!}{
  \begin{tabular}{ccccccccccccc}
   \toprule
    & & & & \multicolumn{9}{c}{$\tau$} \\
    \cmidrule(r){5-13}
    Task & $\lambda$ & Value & Architecture & 0.1 & 0.2 & 0.3 & 0.4 & 0.5 & 0.6 & 0.7 & 0.8 & 0.9 \\ 
    \midrule
    \multirow{6}{*}{\textbf{Nuclei}} & \multirow{6}{*}{\textbf{$\lambda_1$}} & \multirow{3}{*}{0.1} & \emph{HSD-WBR}(ours) & \textbf{0.818} & \textbf{0.818} & \textbf{0.817} & \textbf{0.816} & \textbf{0.809} & \textbf{0.797} & \underline{0.775} & \underline{0.716} & \underline{0.460} \\
    & & & \emph{HSD}(ours) & \underline{0.815} & \underline{0.815} & \underline{0.815} & \underline{0.813} & \underline{0.807} & \underline{0.795} & 0.773 & 0.714 & 0.444 \\
    & & & \emph{SD} (RI)\citep{Stardist18} & 0.811 & 0.811 & 0.811 & 0.809 & 0.804 & \underline{0.795} & \textbf{0.778} & \textbf{0.730} & \textbf{0.501} \\
    \cmidrule(r){3-13}
    & & \multirow{3}{*}{0.4} & \emph{HSD-WBR}(ours) & 0.817 & 0.817 & 0.817 & 0.815 & 0.808 & 0.796 & 0.773 & 0.712 & 0.448 \\
    & & & \emph{HSD}(ours) & \textbf{0.821} & \textbf{0.821} & \textbf{0.820} & \textbf{0.819} & \textbf{0.813} & \underline{0.802} & \underline{0.781} & \underline{0.719} & \underline{0.452} \\
    & & & \emph{SD} (RI)\citep{Stardist18} & \underline{0.819} & \underline{0.819} & \underline{0.819} & \underline{0.818} & \textbf{0.813} & \textbf{0.803} & \textbf{0.785} & \textbf{0.737} & \textbf{0.507}\\
    \midrule
    \multirow{6}{*}{\textbf{Cytoplasm}} & \multirow{6}{*}{\textbf{$\lambda_1$}} & \multirow{3}{*}{0.1} & \emph{HSD-WBR}(ours) & \underline{0.651} & \underline{0.650} & \underline{0.646} & \underline{0.631} & \underline{0.595} & \underline{0.530} & \underline{0.425} & 0.258 & 0.040 \\
    & & & \emph{HSD}(ours) & \textbf{0.662} & \textbf{0.661} & \textbf{0.657} & \textbf{0.642} & \textbf{0.604} & \textbf{0.538} & \textbf{0.432} & \textbf{0.264} & 0.031 \\
    & & & \emph{SD} (RI)\citep{Stardist18} & 0.609 & 0.609 & 0.605 & 0.593 & 0.565 & 0.509 & 0.419 & \underline{0.271} & \textbf{0.060} \\
    \cmidrule(r){3-13}
    & & \multirow{3}{*}{0.4} & \emph{HSD-WBR}(ours) & \underline{0.649} & \underline{0.648} & \underline{0.644} & \underline{0.629} & \underline{0.594} & \underline{0.531} & 0.429 & 0.273 & \underline{0.055} \\
    & & & \emph{HSD}(ours) & \textbf{0.662} & \textbf{0.662} & \textbf{0.658} & \textbf{0.643} & \textbf{0.606} & \textbf{0.544} & \textbf{0.439} & \underline{0.275} & 0.047 \\
    & & & \emph{SD} (RI)\citep{Stardist18} & 0.637 & 0.636 & 0.633 & 0.620 & 0.588 & 0.529 & \underline{0.434} & \textbf{0.283} & \textbf{0.065} \\
  \bottomrule
  \multicolumn{13}{p{350pt}}{\emph{SD} (FT) experiments were omitted since they were pretrained with default $\lambda$s. For those experiments please see tables in the main text.}
  \end{tabular}
  }
\end{table}

\begin{table}[h!]
  \caption{HeLaCytoNuc \emph{AP} (Equation \ref{eq:ap}) at different thresholds ($\tau$) and $\lambda_1$.}
  \label{tab:hela_ablation_lambda1_ap_performance}
  \centering
  \resizebox{\linewidth}{!}{
  \begin{tabular}{ccccccccccccc}
   \toprule
    & & & & \multicolumn{9}{c}{$\tau$} \\
    \cmidrule(r){5-13}
    Task & $\lambda$ & Value & Architecture & 0.1 & 0.2 & 0.3 & 0.4 & 0.5 & 0.6 & 0.7 & 0.8 & 0.9 \\
    \midrule
    \multirow{6}{*}{\textbf{Nuclei}} & \multirow{6}{*}{\textbf{$\lambda_1$}} & \multirow{3}{*}{0.1} & \emph{HSD-WBR}(ours) & \textbf{0.860} & \underline{0.859} & \underline{0.857} & \underline{0.850} & \underline{0.827} & \underline{0.792} & \underline{0.741} & 0.631 & \underline{0.318} \\
    & & & \emph{HSD}(ours) & \underline{0.859} & \underline{0.859} & \underline{0.857} & 0.848 & \underline{0.827} & 0.791 & \underline{0.741} & \underline{0.633} & 0.305 \\
    & & & \emph{SD} (RI)\citep{Stardist18} & \textbf{0.860} & \textbf{0.860} & \textbf{0.858} & \textbf{0.851} & \textbf{0.833} & \textbf{0.805} & \textbf{0.764} & \textbf{0.670} & \textbf{0.367} \\
    \cmidrule(r){3-13}
    & & \multirow{3}{*}{0.4} & \emph{HSD-WBR}(ours) & \underline{0.858} & \underline{0.858} & \underline{0.855} & 0.846 & 0.822 & 0.787 & 0.735 & 0.623 & 0.307 \\
    & & & \emph{HSD}(ours) & \textbf{0.866} & \textbf{0.866} & \textbf{0.864} & \textbf{0.857} & \textbf{0.836} & \textbf{0.804} & \underline{0.753} & \underline{0.639} & \underline{0.311} \\
    & & & \emph{SD} (RI)\citep{Stardist18} & 0.856 & 0.856 & 0.854 & \underline{0.848} & \underline{0.829} & \underline{0.801} & \textbf{0.758} & \textbf{0.666} & \textbf{0.366} \\ 
    \midrule
    \multirow{6}{*}{\textbf{Cytoplasm}} & \multirow{6}{*}{\textbf{$\lambda_1$}} & \multirow{3}{*}{0.1} & \emph{HSD-WBR}(ours) & \underline{0.788} & \underline{0.780} & \underline{0.756} & \underline{0.697} & \underline{0.597} & \underline{0.467} & \underline{0.322} & \underline{0.163} & \underline{0.021} \\
    & & & \emph{HSD}(ours) & 0.777 & 0.768 & 0.744 & 0.686 & 0.585 & 0.460 & 0.318 & \underline{0.163} & 0.016 \\
    & & & \emph{SD} (RI)\citep{Stardist18} & \textbf{0.794} & \textbf{0.788} & \textbf{0.766} & \textbf{0.715} & \textbf{0.624} & \textbf{0.499} & \textbf{0.355} & \textbf{0.192} & \textbf{0.035} \\
    \cmidrule(r){3-13}
    & & \multirow{3}{*}{0.4} & \emph{HSD-WBR}(ours) & 0.779 & 0.771 & 0.746 & 0.689 & 0.591 & 0.467 & 0.325 & \underline{0.174} & \underline{0.029} \\
    & & & \emph{HSD}(ours) & \underline{0.796} & \underline{0.788} & \underline{0.765} & \underline{0.708} & \underline{0.603} & \underline{0.480} & \underline{0.333} & \underline{0.174} & 0.024 \\
    & & & \emph{SD} (RI)\citep{Stardist18} & \textbf{0.802} & \textbf{0.796} & \textbf{0.774} & \textbf{0.719} & \textbf{0.623} & \textbf{0.496} & \textbf{0.350} & \textbf{0.192} & \textbf{0.036} \\
  \bottomrule
  \multicolumn{13}{p{350pt}}{\emph{SD} (FT) experiments were omitted since they were pretrained with default $\lambda$s. For those experiments please see tables in the main text.}
  \end{tabular}
  }
\end{table}

\begin{table}[h!]
  \caption{HeLaCytoNuc \emph{$IoU_R$} (Equation \ref{eq:iou-recall}) at different thresholds ($\tau$) and $\lambda_2$.}
  \label{tab:hela_ablation_lambda2_iou_performance}
  \centering
  \resizebox{\linewidth}{!}{
  \begin{tabular}{ccccccccccccc}
   \toprule
    & & & & \multicolumn{9}{c}{$\tau$} \\
    \cmidrule(r){5-13}
    Task & $\lambda$ & Value & Architecture & 0.1 & 0.2 & 0.3 & 0.4 & 0.5 & 0.6 & 0.7 & 0.8 & 0.9 \\
    \midrule
    \multirow{6}{*}{\textbf{Nuclei}} & \multirow{6}{*}{\textbf{$\lambda_2$}} & \multirow{3}{*}{5e-5} & \emph{HSD-WBR}(ours) & \textbf{0.825} & \textbf{0.825} & \textbf{0.825} & \textbf{0.823} & \textbf{0.818} & \textbf{0.806} & \textbf{0.784} & \textbf{0.722} & \textbf{0.462}\\
    & & & \emph{HSD}(ours) & \underline{0.817} & \underline{0.817} & \underline{0.816} & \underline{0.815} & \underline{0.809} & \underline{0.796} & \underline{0.775} & 0.714 & 0.458 \\
    & & & \emph{SD} (RI)\citep{Stardist18} & 0.801 & 0.801 & 0.800 & 0.799 & 0.795 & 0.785 & 0.769 & \underline{0.721} & \underline{0.501} \\
    \cmidrule(r){3-13}
    & & \multirow{3}{*}{2e-4} & \emph{HSD-WBR}(ours) & \underline{0.827} & \underline{0.827} & \underline{0.826} & \underline{0.825} & \underline{0.819} & \underline{0.807} & \underline{0.785} & 0.725 & 0.467 \\
    & & & \emph{HSD}(ours) & \textbf{0.833} & \textbf{0.833} & \textbf{0.833} & \textbf{0.831} & \textbf{0.823} & \textbf{0.813} & \textbf{0.790} & \underline{0.729} & \underline{0.470} \\
    & & & \emph{SD} (RI)\citep{Stardist18} & 0.810 & 0.810 & 0.810 & 0.809 & 0.804 & 0.795 & 0.776 & \textbf{0.730} & \textbf{0.505} \\
    \midrule
    \multirow{6}{*}{\textbf{Cytoplasm}} & \multirow{6}{*}{\textbf{$\lambda_2$}} & \multirow{3}{*}{5e-5} & \emph{HSD-WBR}(ours) & \textbf{0.667} & \textbf{0.666} & \textbf{0.662} & \textbf{0.647} & \textbf{0.612} & \textbf{0.546} & \textbf{0.444} & \textbf{0.285} & \underline{0.058} \\
    & & & \emph{HSD}(ours) & \underline{0.653} & \underline{0.652} & \underline{0.648} & \underline{0.634} & \underline{0.598} & \underline{0.533} & \underline{0.426} & 0.259 & 0.043 \\
    & & & \emph{SD} (RI)\citep{Stardist18} & 0.615 & 0.615 & 0.611 & 0.600 & 0.570 & 0.515 & 0.423 & \underline{0.273} & \textbf{0.060} \\
    \cmidrule(r){3-13}
    & & \multirow{3}{*}{2e-4} & \emph{HSD-WBR}(ours) & \textbf{0.660} & \textbf{0.660} & \textbf{0.655} & \underline{0.640} & \underline{0.603} & \underline{0.538} & \underline{0.433} & \underline{0.272} & \textbf{0.060} \\
    & & & \emph{HSD}(ours) & \underline{0.660} & \underline{0.659} & \underline{0.655} & \textbf{0.641} & \textbf{0.605} & \textbf{0.542} & \textbf{0.438} & \textbf{0.279} & \underline{0.057} \\
    & & & \emph{SD} (RI)\citep{Stardist18} & 0.631 & 0.631 & 0.627 & 0.612 & 0.578 & 0.518 & 0.416 & 0.259 & 0.050 \\
  \bottomrule
  \multicolumn{13}{p{350pt}}{\emph{SD} (FT) experiments were omitted since they were pretrained with default $\lambda$s. For those experiments please see tables in the main text.}
  \end{tabular}
  }
\end{table}

\begin{table}[h!]
  \caption{HeLaCytoNuc \emph{AP} (Equation \ref{eq:ap}) at different thresholds ($\tau$) and $\lambda_2$.}
  \label{tab:hela_ablation_lambda2_ap_performance}
  \centering
  \resizebox{\linewidth}{!}{
  \begin{tabular}{ccccccccccccc}
   \toprule
    & & & & \multicolumn{9}{c}{$\tau$} \\
    \cmidrule(r){5-13}
    Task & $\lambda$ & Value & Architecture & 0.1 & 0.2 & 0.3 & 0.4 & 0.5 & 0.6 & 0.7 & 0.8 & 0.9 \\
    \midrule
    \multirow{6}{*}{\textbf{Nuclei}} & \multirow{6}{*}{\textbf{$\lambda_2$}} & \multirow{3}{*}{5e-5} & \emph{HSD-WBR}(ours) & \textbf{0.864} & \textbf{0.864} & \textbf{0.862} & \textbf{0.854} & \textbf{0.835} & \textbf{0.799} & \underline{0.748} & \underline{0.635} & \underline{0.318} \\
    & & & \emph{HSD}(ours) & \underline{0.855} & \underline{0.855} & \underline{0.853} & \underline{0.844} & \underline{0.823} & 0.788 & 0.738 & 0.627 & 0.315 \\
    & & & \emph{SD} (RI)\citep{Stardist18} & 0.849 & 0.849 & 0.847 & 0.841 & 0.823 & \underline{0.798} & \textbf{0.757} & \textbf{0.664} & \textbf{0.370} \\
    \cmidrule(r){3-13}
    & & \multirow{3}{*}{2e-4} & \emph{HSD-WBR}(ours) & \underline{0.859} & \underline{0.859} & \underline{0.857} & \underline{0.849} & 0.828 & 0.793 & 0.742 & \underline{0.633} & \underline{0.320} \\
    & & & \emph{HSD}(ours) & \textbf{0.864} & \textbf{0.864} & \textbf{0.861} & \textbf{0.853} & \textbf{0.830} & \underline{0.796} & \underline{0.744} & \underline{0.633} & \underline{0.320} \\
    & & & \emph{SD} (RI)\citep{Stardist18} & 0.854 & 0.854 & 0.852 & 0.846 & \underline{0.829} & \textbf{0.802} & \textbf{0.757} & \textbf{0.666} & \textbf{0.369} \\ 
    \midrule
    \multirow{6}{*}{\textbf{Cytoplasm}} & \multirow{6}{*}{\textbf{$\lambda_2$}} & \multirow{3}{*}{5e-5} & \emph{HSD-WBR}(ours) & 0.779 & 0.769 & 0.747 & 0.689 & 0.593 & 0.467 & \underline{0.327} & \underline{0.177} & \underline{0.030} \\
    & & & \emph{HSD}(ours) & \underline{0.785} & \underline{0.777} & \underline{0.755} & \underline{0.697} & \underline{0.599} & \underline{0.469} & 0.322 & 0.163 & 0.022 \\
    & & & \emph{SD} (RI)\citep{Stardist18} & \textbf{0.797} & \textbf{0.791} & \textbf{0.771} & \textbf{0.720} & \textbf{0.627} & \textbf{0.503} & \textbf{0.357} & \textbf{0.192} & \textbf{0.034} \\
    \cmidrule(r){3-13}
    & & \multirow{3}{*}{2e-4} & \emph{HSD-WBR}(ours) & 0.781 & 0.773 & 0.748 & 0.691 & 0.590 & 0.464 & 0.321 & 0.169 & \textbf{0.031} \\
    & & & \emph{HSD}(ours) & \underline{0.788} & \underline{0.781} & \underline{0.758} & \underline{0.702} & \underline{0.603} & \underline{0.477} & \underline{0.331} & \textbf{0.177} & \underline{0.030} \\
    & & & \emph{SD} (RI)\citep{Stardist18} & \textbf{0.804} & \textbf{0.797} & \textbf{0.773} & \textbf{0.713} & \textbf{0.612} & \textbf{0.484} & \textbf{0.333} & \underline{0.173} & 0.027 \\
  \bottomrule
  \multicolumn{13}{p{350pt}}{\emph{SD} (FT) experiments were omitted since they were pretrained with default $\lambda$s. For those experiments please see tables in the main text.}
  \end{tabular}
  }
\end{table}

\begin{table}[h!]
  \caption{HeLaCytoNuc \emph{$IoU_R$} (Equation \ref{eq:iou-recall}) at different thresholds ($\tau$) and $\lambda_3$.}
  \label{tab:hela_ablation_lambda3_iou_performance}
  \centering
  \resizebox{\linewidth}{!}{
  \begin{tabular}{ccccccccccccc}
   \toprule
    & & & & \multicolumn{9}{c}{$\tau$} \\
    \cmidrule(r){5-13}
    Task & $\lambda$ & Value & Architecture & 0.1 & 0.2 & 0.3 & 0.4 & 0.5 & 0.6 & 0.7 & 0.8 & 0.9 \\
    \midrule
    \multirow{2}{*}{\textbf{Nuclei}} & \multirow{2}{*}{\textbf{$\lambda_3$}} & \multirow{1}{*}{0.5} & \multirow{2}{*}{\emph{HSD-WBR}(ours)} & 0.818 & 0.818 & 0.818 & 0.816 & 0.811 & 0.800 & 0.777 & 0.705 & 0.402 \\
    & & \multirow{1}{*}{2} & & \textbf{0.830} & \textbf{0.830} & \textbf{0.830} & \textbf{0.828} & \textbf{0.822} & \textbf{0.810} & \textbf{0.788} & \textbf{0.729} & \textbf{0.472} \\
    \midrule
    \multirow{2}{*}{\textbf{Cytoplasm}} & \multirow{2}{*}{\textbf{$\lambda_3$}} & \multirow{1}{*}{0.5} & \multirow{2}{*}{\emph{HSD-WBR}(ours)} & 0.661 & 0.660 & 0.656 & 0.641 & 0.605 & 0.541 & 0.439 & 0.279 & \textbf{0.062} \\
    & & \multirow{1}{*}{2} & & \textbf{0.670} & \textbf{0.669} & \textbf{0.665} & \textbf{0.650} & \textbf{0.615} & \textbf{0.550} & \textbf{0.444} & \textbf{0.282} & 0.060 \\
  \bottomrule
  \multicolumn{13}{p{350pt}}{Being the only architecture using $\lambda_3$, only HSD-WBR is considered.}
  \end{tabular}
  }
\end{table}

\begin{table}[h!]
  \caption{HeLaCytoNuc \emph{AP} (Equation \ref{eq:ap}) at different thresholds ($\tau$) and $\lambda_3$.}
  \label{tab:hela_ablation_lambda3_ap_performance}
  \centering
  \resizebox{\linewidth}{!}{
  \begin{tabular}{ccccccccccccc}
   \toprule
    & & & & \multicolumn{9}{c}{$\tau$} \\
    \cmidrule(r){5-13}
    Task & $\lambda$ & Value & Architecture & 0.1 & 0.2 & 0.3 & 0.4 & 0.5 & 0.6 & 0.7 & 0.8 & 0.9 \\
    \midrule
    \multirow{2}{*}{\textbf{Nuclei}} & \multirow{2}{*}{\textbf{$\lambda_3$}} & \multirow{1}{*}{0.5} & \multirow{2}{*}{\emph{HSD-WBR}(ours)} & 0.864 & 0.864 & 0.862 & \textbf{0.856} & \textbf{0.835} & \textbf{0.804} & \textbf{0.750} & 0.618 & 0.266 \\
    & & \multirow{1}{*}{2} & & \textbf{0.866} & \textbf{0.865} & \textbf{0.863} & 0.855 & 0.834 & 0.798 & 0.747 & \textbf{0.638} & \textbf{0.324} \\
    \midrule
    \multirow{2}{*}{\textbf{Cytoplasm}} & \multirow{2}{*}{\textbf{$\lambda_3$}} & \multirow{1}{*}{0.5} & \multirow{2}{*}{\emph{HSD-WBR}(ours)} & 0.780 & 0.772 & 0.748 & 0.689 & 0.591 & 0.467 & 0.328 & 0.174 & \textbf{0.032} \\
    & & \multirow{1}{*}{2} & & \textbf{0.788} & \textbf{0.781} & \textbf{0.758} & \textbf{0.700} & \textbf{0.603} & \textbf{0.476} & \textbf{0.330} & \textbf{0.176} & 0.031 \\
  \bottomrule
  \multicolumn{13}{p{350pt}}{Being the only architecture using $\lambda_3$, only HSD-WBR is considered.}
  \end{tabular}
  }
\end{table}

\begin{table}[h!]
  \caption{HeLaCytoNuc \emph{JTPR} (Equations \ref{eq:jtp_rate_inner}-\ref{eq:jtp_rate_outer}) at different $\lambda_1$.}
  \label{tab:hela_ablation_lambda1_joint_tp_rate_performance}
  \centering
  \resizebox{0.75\linewidth}{!}{
  \begin{tabular}{ccccc}
   \toprule
    & & & inner nested & outer nested \\
    $\lambda$ & Value & Architecture & object $\uparrow$ & object $\uparrow$ \\
    & & & (\emph{Nuclei}) & (\emph{Cytoplasm}) \\
    \midrule
    \multirow{6}{*}{\textbf{$\lambda_1$}} & \multirow{3}{*}{0.1} & \emph{HSD-WBR}(ours) & \underline{0.886} & \underline{0.843} \\
    & & \emph{HSD}(ours) & \textbf{0.891} & \textbf{0.848} \\
    & & \emph{SD} (RI)\citep{Stardist18} & 0.798 & 0.751 \\
    \cmidrule(r){2-5}
    & \multirow{3}{*}{0.4} & \emph{HSD-WBR}(ours) & \textbf{0.890} & \textbf{0.848} \\
    & & \emph{HSD}(ours) & \underline{0.884} & \underline{0.839} \\
    & & \emph{SD} (RI)\citep{Stardist18} & 0.834 & 0.788 \\
  \bottomrule
  \multicolumn{5}{p{300pt}}{\emph{SD} (FT) experiments were omitted since they were pretrained with default $\lambda$s. For those experiments please see tables in the main text.}
  \end{tabular}
  }
\end{table}

\begin{table}[h!]
  \caption{HeLaCytoNuc \emph{JTPR} (Equations \ref{eq:jtp_rate_inner}-\ref{eq:jtp_rate_outer}) at different $\lambda_2$.}
  \label{tab:hela_ablation_lambda2_joint_tp_rate_performance}
  \centering
  \resizebox{0.75\linewidth}{!}{
  \begin{tabular}{ccccc}
   \toprule
    & & & inner nested & outer nested \\
    $\lambda$ & Value & Architecture & object $\uparrow$ & object $\uparrow$ \\
    & & & (\emph{Nuclei}) & (\emph{Cytoplasm}) \\
    \midrule
    \multirow{6}{*}{\textbf{$\lambda_2$}} & \multirow{3}{*}{5e-5} & \emph{HSD-WBR}(ours) & \textbf{0.898} & \textbf{0.853} \\
    & & \emph{HSD}(ours) & \underline{0.889} & \underline{0.846} \\
    & & \emph{SD} (RI)\citep{Stardist18} & 0.801 &  0.755 \\
    \cmidrule(r){2-5}
    & \multirow{3}{*}{2e-4} & \emph{HSD-WBR}(ours) & \textbf{0.898} & \textbf{0.855} \\
    & & \emph{HSD}(ours) & \underline{0.897} &  \textbf{0.855} \\
    & & \emph{SD} (RI)\citep{Stardist18} & 0.836 & 0.788 \\
  \bottomrule
  \multicolumn{5}{p{300pt}}{\emph{SD} (FT) experiments were omitted since they were pretrained with default $\lambda$s. For those experiments please see tables in the main text.}
  \end{tabular}
  }
\end{table}

\begin{table}[h!]
  \caption{HeLaCytoNuc \emph{JTPR} (Equations \ref{eq:jtp_rate_inner}-\ref{eq:jtp_rate_outer}) at different $\lambda_3$.}
  \label{tab:hela_ablation_lambda3_joint_tp_rate_performance}
  \centering
  \resizebox{0.75\linewidth}{!}{
  \begin{tabular}{ccccc}
   \toprule
    & & & inner nested & outer nested \\
    $\lambda$ & Value & Architecture & object $\uparrow$ & object $\uparrow$ \\
    & & & (\emph{Nuclei}) & (\emph{Cytoplasm}) \\
    \midrule
    \multirow{2}{*}{\textbf{$\lambda_3$}} & \multirow{1}{*}{0.5} & \multirow{2}{*}{\emph{HSD-WBR}(ours)} & 0.893 & 0.848 \\
    & 2 & & \textbf{0.901} & \textbf{0.859} \\
  \bottomrule
  \multicolumn{5}{p{300pt}}{Being the only architecture using $\lambda_3$, only HSD-WBR is considered.}
  \end{tabular}
  }
\end{table}

\begin{table}[h!]
  \caption{VACVPlaque \emph{$IoU_R$} (Equation \ref{eq:iou-recall}) at different thresholds ($\tau$) and $\lambda_1$.}
  \label{tab:vacv_ablation_lambda1_iou_performance}
  \centering
  \resizebox{\linewidth}{!}{
  \begin{tabular}{ccccccccccccc}
   \toprule
    & & & & \multicolumn{9}{c}{$\tau$} \\
    \cmidrule(r){5-13}
    Task & $\lambda$ & Value & Architecture & 0.1 & 0.2 & 0.3 & 0.4 & 0.5 & 0.6 & 0.7 & 0.8 & 0.9 \\ 
    \midrule
    \multirow{6}{*}{\textbf{Plaques}} & \multirow{6}{*}{\textbf{$\lambda_1$}} & \multirow{3}{*}{0.1} & \emph{HSD-WBR}(ours) & \underline{0.408} & \underline{0.407} & \underline{0.400} & \underline{0.384} & \underline{0.355} & \underline{0.284} & \underline{0.174} & \underline{0.068} & \underline{0.010} \\
    & & & \emph{HSD}(ours) & 0.384 & 0.383 & 0.377 & 0.362 & \underline{0.335} & 0.265 & 0.156 & 0.060 & \underline{0.010} \\
    & & & \emph{SD} (RI)\citep{Stardist18} & \textbf{0.495} & \textbf{0.494} & \textbf{0.488} & \textbf{0.474} & \textbf{0.445} & \textbf{0.364} & \textbf{0.238} & \textbf{0.099} & \textbf{0.018} \\
    \cmidrule(r){3-13}
    & & \multirow{3}{*}{0.4} & \emph{HSD-WBR}(ours) & \underline{0.398} & \underline{0.397} & \underline{0.389} & \underline{0.372} & \underline{0.342} & \underline{0.272} & \underline{0.164} & 0.063 & 0.009 \\
    & & & \emph{HSD}(ours) & 0.387 & 0.385 & 0.376 & 0.357 & 0.324 & 0.256 & 0.160 & \underline{0.067} & \underline{0.011} \\
    & & & \emph{SD} (RI)\citep{Stardist18} & \textbf{0.491} & \textbf{0.490} & \textbf{0.485} & \textbf{0.473} & \textbf{0.445} & \textbf{0.375} & \textbf{0.253} & \textbf{0.116} & \textbf{0.021} \\
    \midrule
    \multirow{6}{*}{\textbf{Wells}} & \multirow{6}{*}{\textbf{$\lambda_1$}} & \multirow{3}{*}{0.1} & \emph{HSD-WBR}(ours) & \underline{0.949} & \underline{0.949} & \underline{0.949} & \underline{0.949} & \underline{0.949} & \underline{0.949} & \underline{0.943} & \underline{0.943} & \underline{0.943} \\
    & & & \emph{HSD}(ours) & \textbf{0.956} & \textbf{0.956} & \textbf{0.956} & \textbf{0.956} & \textbf{0.956} & \textbf{0.956} & \textbf{0.956} & \textbf{0.956} & \textbf{0.947} \\
    & & & \emph{SD} (RI)\citep{Stardist18} & 0.943 & 0.943 & 0.943 & 0.943 & 0.943 & 0.943 & 0.943 & 0.937 & 0.930 \\
    \cmidrule(r){3-13}
    & & \multirow{3}{*}{0.4} & \emph{HSD-WBR}(ours) & \textbf{0.957} & \textbf{0.957} & \textbf{0.957} & \textbf{0.957} & \textbf{0.957} & \textbf{0.957} & \textbf{0.957} & \textbf{0.957} & \textbf{0.957} \\
    & & & \emph{HSD}(ours) & \underline{0.956} & \underline{0.956} & \underline{0.956} & \underline{0.956} & \underline{0.956} & \underline{0.956} & \underline{0.956} & \underline{0.956} & \underline{0.948} \\
    & & & \emph{SD} (RI)\citep{Stardist18} & 0.942 & 0.942 & 0.942 & 0.942 & 0.942 & 0.942 & 0.942 & 0.942 & 0.915 \\
  \bottomrule
  \multicolumn{13}{p{350pt}}{\emph{SD} (FT) experiments were omitted since they were pretrained with default $\lambda$s. For those experiments please see tables in the main text.}
  \end{tabular}
  }
\end{table}

\begin{table}[h!]
  \caption{VACVPlaque \emph{AP} (Equation \ref{eq:ap}) at different thresholds ($\tau$) and $\lambda_1$.}
  \label{tab:vacv_ablation_lambda1_ap_performance}
  \centering
  \resizebox{\linewidth}{!}{
  \begin{tabular}{ccccccccccccc}
   \toprule
    & & & & \multicolumn{9}{c}{$\tau$} \\
    \cmidrule(r){5-13}
    Task & $\lambda$ & Value & Architecture & 0.1 & 0.2 & 0.3 & 0.4 & 0.5 & 0.6 & 0.7 & 0.8 & 0.9 \\
    \midrule
    \multirow{6}{*}{\textbf{Plaques}} & \multirow{6}{*}{\textbf{$\lambda_1$}} & \multirow{3}{*}{0.1} & \emph{HSD-WBR}(ours) & \underline{0.577} & \underline{0.566} & \underline{0.529} & \underline{0.473} & \underline{0.400} & \underline{0.273} & \underline{0.138} & \underline{0.046} & \underline{0.006} \\
    & & & \emph{HSD}(ours) & 0.550 & 0.539 & 0.508 & 0.455 & 0.386 & 0.261 & 0.126 & 0.041 & \underline{0.006} \\
    & & & \emph{SD} (RI)\citep{Stardist18} & \textbf{0.737} & \textbf{0.729} & \textbf{0.693} & \textbf{0.632} & \textbf{0.541} & \textbf{0.370} & \textbf{0.197} & \textbf{0.068} & \textbf{0.010} \\
    \cmidrule(r){3-13}
    & & \multirow{3}{*}{0.4} & \emph{HSD-WBR}(ours) & \underline{0.533} & \underline{0.521} & \underline{0.486} & \underline{0.430} & \underline{0.361} & \underline{0.247} & \underline{0.124} & 0.040 & 0.005 \\
    & & & \emph{HSD}(ours) & 0.518 & 0.503 & 0.464 & 0.405 & 0.332 & 0.226 & 0.119 & \underline{0.043} & \underline{0.006} \\
    & & & \emph{SD} (RI)\citep{Stardist18} & \textbf{0.713} & \textbf{0.707} & \textbf{0.675} & \textbf{0.622} & \textbf{0.537} & \textbf{0.385} & \textbf{0.213} & \textbf{0.080} & \textbf{0.013} \\ 
    \midrule
    \multirow{6}{*}{\textbf{Wells}} & \multirow{6}{*}{\textbf{$\lambda_1$}} & \multirow{3}{*}{0.1} & \emph{HSD-WBR}(ours) & \underline{0.817} & \underline{0.817} & \underline{0.817} & \underline{0.817} & \underline{0.817} & \underline{0.817} & 0.803 & 0.803 & \underline{0.803} \\
    & & & \emph{HSD}(ours) & \underline{0.817} & \underline{0.817} & \underline{0.817} & \underline{0.817} & \underline{0.817} & \underline{0.817} & \underline{0.817} & \underline{0.817} & 0.803 \\
    & & & \emph{SD} (RI)\citep{Stardist18} & \textbf{0.992} & \textbf{0.992} & \textbf{0.992} & \textbf{0.992} & \textbf{0.992} & \textbf{0.992} & \textbf{0.992} & \textbf{0.976} & \textbf{0.961} \\
    \cmidrule(r){3-13}
    & & \multirow{3}{*}{0.4} & \emph{HSD-WBR}(ours) & 0.763 & 0.763 & 0.763 & 0.763 &0.763 & 0.763 & 0.763 & 0.763 & 0.763 \\
    & & & \emph{HSD}(ours) & \underline{0.824} & \underline{0.824} & \underline{0.824} & \underline{0.824} & \underline{0.824} & \underline{0.824} & \underline{0.824} & \underline{0.824} & \underline{0.810} \\
    & & & \emph{SD} (RI)\citep{Stardist18} & \textbf{0.992} & \textbf{0.992} & \textbf{0.992} & \textbf{0.992} & \textbf{0.992} & \textbf{0.992} & \textbf{0.992} & \textbf{0.992} & \textbf{0.931} \\
  \bottomrule
  \multicolumn{13}{p{350pt}}{\emph{SD} (FT) experiments were omitted since they were pretrained with default $\lambda$s. For those experiments please see tables in the main text.}
  \end{tabular}
  }
\end{table}

\begin{table}[h!]
  \caption{VACVPlaque \emph{$IoU_R$} (Equation \ref{eq:iou-recall}) at different thresholds ($\tau$) and $\lambda_2$.}
  \label{tab:vacv_ablation_lambda2_iou_performance}
  \centering
  \resizebox{\linewidth}{!}{
  \begin{tabular}{ccccccccccccc}
   \toprule
    & & & & \multicolumn{9}{c}{$\tau$} \\
    \cmidrule(r){5-13}
    Task & $\lambda$ & Value & Architecture & 0.1 & 0.2 & 0.3 & 0.4 & 0.5 & 0.6 & 0.7 & 0.8 & 0.9 \\
    \midrule
    \multirow{6}{*}{\textbf{Plaques}} & \multirow{6}{*}{\textbf{$\lambda_2$}} & \multirow{3}{*}{5e-5} & \emph{HSD-WBR}(ours) & 0.398 & 0.396 & 0.388 & 0.370 & 0.337 & 0.265 & \underline{0.161} & \underline{0.067} & \underline{0.009} \\
    & & & \emph{HSD}(ours) & \underline{0.415} & \underline{0.413} & \underline{0.405} & \underline{0.385} & \underline{0.351} & \underline{0.271} & 0.159 & 0.060 & 0.008 \\
    & & & \emph{SD} (RI)\citep{Stardist18} & \textbf{0.501} & \textbf{0.501} & \textbf{0.496} & \textbf{0.482} & \textbf{0.453} & \textbf{0.376} & \textbf{0.251} & \textbf{0.112} & \textbf{0.020} \\
    \cmidrule(r){3-13}
    & & \multirow{3}{*}{2e-4} & \emph{HSD-WBR}(ours) & \underline{0.403} & \underline{0.401} & \underline{0.391} & \underline{0.371} & \underline{0.335} & \underline{0.259} & \underline{0.157} & \underline{0.061} & \underline{0.010} \\
    & & & \emph{HSD}(ours) & 0.358 & 0.356 & 0.349 & 0.331 & 0.298 & 0.227 & 0.132 & 0.051 & 0.007 \\
    & & & \emph{SD} (RI)\citep{Stardist18} & \textbf{0.508} & \textbf{0.508} & \textbf{0.502} & \textbf{0.488} & \textbf{0.460} & \textbf{0.381} & \textbf{0.258} & \textbf{0.118} & \textbf{0.022} \\
    \midrule
    \multirow{6}{*}{\textbf{Wells}} & \multirow{6}{*}{\textbf{$\lambda_2$}} & \multirow{3}{*}{5e-5} & \emph{HSD-WBR}(ours) & \underline{0.952} & \underline{0.952} & \underline{0.952} & \underline{0.952} & \underline{0.952} & \underline{0.952} & \underline{0.952} & \underline{0.952} & \underline{0.943} \\
    & & & \emph{HSD}(ours) & \textbf{0.958} & \textbf{0.958} & \textbf{0.958} & \textbf{0.958} & \textbf{0.958} & \textbf{0.958} & \textbf{0.958} & \textbf{0.958} & \textbf{0.958} \\
    & & & \emph{SD} (RI)\citep{Stardist18} & 0.942 & 0.942 & 0.942 & 0.942 & 0.942 & 0.942 & 0.937 & 0.937 & 0.930 \\
    \cmidrule(r){3-13}
    & & \multirow{3}{*}{2e-4} & \emph{HSD-WBR}(ours) & \textbf{0.953} & \textbf{0.953} & \textbf{0.953} & \textbf{0.953} & \textbf{0.953} & \textbf{0.953} & \textbf{0.953} & \textbf{0.953} & \textbf{0.953} \\
    & & & \emph{HSD}(ours) & 0.941 & 0.941 & 0.941 & 0.941 & 0.941 & 0.941 & 0.941 & 0.933 & \underline{0.908} \\
    & & & \emph{SD} (RI)\citep{Stardist18} & \underline{0.947} & \underline{0.947} & \underline{0.947} & \underline{0.947} & \underline{0.947} & \underline{0.947} & \underline{0.942} & \underline{0.942} & \underline{0.908} \\
  \bottomrule
  \multicolumn{13}{p{350pt}}{\emph{SD} (FT) experiments were omitted since they were pretrained with default $\lambda$s. For those experiments please see tables in the main text.}
  \end{tabular}
  }
\end{table}

\begin{table}[h!]
  \caption{VACVPlaque \emph{AP} (Equation \ref{eq:ap}) at different thresholds ($\tau$) and $\lambda_2$.}
  \label{tab:vacv_ablation_lambda2_ap_performance}
  \centering
  \resizebox{\linewidth}{!}{
  \begin{tabular}{ccccccccccccc}
   \toprule
    & & & & \multicolumn{9}{c}{$\tau$} \\
    \cmidrule(r){5-13}
    Task & $\lambda$ & Value & Architecture & 0.1 & 0.2 & 0.3 & 0.4 & 0.5 & 0.6 & 0.7 & 0.8 & 0.9 \\
    \midrule
    \multirow{6}{*}{\textbf{Plaques}} & \multirow{6}{*}{\textbf{$\lambda_2$}} & \multirow{3}{*}{5e-5} & \emph{HSD-WBR}(ours) & 0.539 & 0.523 & 0.487 & 0.426 & 0.351 & 0.235 & 0.120 & 0.043 & \underline{0.005} \\
    & & & \emph{HSD}(ours) & \underline{0.571} & \underline{0.555} & \underline{0.512} & \underline{0.448} & \underline{0.370} & \underline{0.241} & \underline{0.117} & \underline{0.038} & \underline{0.005} \\
    & & & \emph{SD} (RI)\citep{Stardist18} & \textbf{0.725} & \textbf{0.718} & \textbf{0.688} & \textbf{0.630} & \textbf{0.541} & \textbf{0.379} & \textbf{0.207} & \textbf{0.076} & \textbf{0.011} \\
    \cmidrule(r){3-13}
    & & \multirow{3}{*}{2e-4} & \emph{HSD-WBR}(ours) & \underline{0.540} & \underline{0.523} & \underline{0.479} & \underline{0.416} & \underline{0.337} & \underline{0.222} & \underline{0.113} & \underline{0.038} & \underline{0.005} \\
    & & & \emph{HSD}(ours) & 0.508 & 0.494 & 0.458 & 0.400 & 0.325 & 0.209 & 0.102 & 0.035 & 0.004 \\
    & & & \emph{SD} (RI)\citep{Stardist18} & \textbf{0.740} & \textbf{0.734} & \textbf{0.700} & \textbf{0.640} & \textbf{0.552} & \textbf{0.383} & \textbf{0.213} & \textbf{0.080} & \textbf{0.012} \\ 
    \midrule
    \multirow{6}{*}{\textbf{Wells}} & \multirow{6}{*}{\textbf{$\lambda_2$}} & \multirow{3}{*}{5e-5} & \emph{HSD-WBR}(ours) & \underline{0.824} & \underline{0.824} & \underline{0.824} & \underline{0.824} & \underline{0.824} & \underline{0.824} & \underline{0.824} & \underline{0.824} & 0.810 \\
    & & & \emph{HSD}(ours) & \underline{0.824} & \underline{0.824} & \underline{0.824} & \underline{0.824} & \underline{0.824} & \underline{0.824} & \underline{0.824} & \underline{0.824} & \underline{0.824} \\
    & & & \emph{SD} (RI)\citep{Stardist18} & \textbf{0.992} & \textbf{0.992} & \textbf{0.992} & \textbf{0.992} & \textbf{0.992} & \textbf{0.992} & \textbf{0.976} & \textbf{0.976} & \textbf{0.961} \\
    \cmidrule(r){3-13}
    & & \multirow{3}{*}{2e-4} & \emph{HSD-WBR}(ours) & \underline{0.824} & \underline{0.824} & \underline{0.824} & \underline{0.824} & \underline{0.824} & \underline{0.824} & \underline{0.824} & \underline{0.824} & \underline{0.824} \\
    & & & \emph{HSD}(ours) & \underline{0.824} & \underline{0.824} & \underline{0.824} & \underline{0.824} & \underline{0.824} & \underline{0.824} & \underline{0.824} & 0.810 & 0.767 \\
    & & & \emph{SD} (RI)\citep{Stardist18} & \textbf{1.000} & \textbf{1.000} & \textbf{1.000} & \textbf{1.000} & \textbf{1.000} & \textbf{1.000} & \textbf{0.984} & \textbf{0.984} & \textbf{0.909} \\
  \bottomrule
  \multicolumn{13}{p{350pt}}{\emph{SD} (FT) experiments were omitted since they were pretrained with default $\lambda$s. For those experiments please see tables in the main text.}
  \end{tabular}
  }
\end{table}

\begin{table}[h!]
  \caption{VACVPlaque \emph{$IoU_R$} (Equation \ref{eq:iou-recall}) at different thresholds ($\tau$) and $\lambda_3$.}
  \label{tab:vacv_ablation_lambda3_iou_performance}
  \centering
  \resizebox{\linewidth}{!}{
  \begin{tabular}{ccccccccccccc}
   \toprule
    & & & & \multicolumn{9}{c}{$\tau$} \\
    \cmidrule(r){5-13}
    Task & $\lambda$ & Value & Architecture & 0.1 & 0.2 & 0.3 & 0.4 & 0.5 & 0.6 & 0.7 & 0.8 & 0.9 \\
    \midrule
    \multirow{2}{*}{\textbf{Plaques}} & \multirow{2}{*}{\textbf{$\lambda_3$}} & \multirow{1}{*}{0.5} & \multirow{2}{*}{\emph{HSD-WBR}(ours)} & \textbf{0.364} & \textbf{0.363} & \textbf{0.355} & 0.337 & 0.305 & \textbf{0.238} & \textbf{0.144} & \textbf{0.055} & \textbf{0.007} \\
    & & \multirow{1}{*}{2} & & 0.363 & 0.361 & 0.354 & \textbf{0.338} & \textbf{0.306} & 0.235 & 0.138 & 0.052 & \textbf{0.007} \\
    \midrule
    \multirow{2}{*}{\textbf{Wells}} & \multirow{2}{*}{\textbf{$\lambda_3$}} & \multirow{1}{*}{0.5} & \multirow{2}{*}{\emph{HSD-WBR}(ours)} & 0.948 & 0.948 & 0.948 & 0.948 & 0.948 & 0.948 & 0.948 & 0.941 & 0.925 \\
    & & \multirow{1}{*}{2} & & \textbf{0.953} & \textbf{0.953} & \textbf{0.953} & \textbf{0.953} & \textbf{0.953} & \textbf{0.953} & \textbf{0.953} & \textbf{0.953} & \textbf{0.953} \\
  \bottomrule
  \multicolumn{13}{p{350pt}}{Being the only architecture using $\lambda_3$, only HSD-WBR is considered.}
  \end{tabular}
  }
\end{table}

\begin{table}[h!]
  \caption{VACVPlaque \emph{AP} (Equation \ref{eq:ap}) at different thresholds ($\tau$) and $\lambda_3$.}
  \label{tab:vacv_ablation_lambda3_ap_performance}
  \centering
  \resizebox{\linewidth}{!}{
  \begin{tabular}{ccccccccccccc}
   \toprule
    & & & & \multicolumn{9}{c}{$\tau$} \\
    \cmidrule(r){5-13}
    Task & $\lambda$ & Value & Architecture & 0.1 & 0.2 & 0.3 & 0.4 & 0.5 & 0.6 & 0.7 & 0.8 & 0.9 \\
    \midrule
    \multirow{2}{*}{\textbf{Plaques}} & \multirow{2}{*}{\textbf{$\lambda_3$}} & \multirow{1}{*}{0.5} & \multirow{2}{*}{\emph{HSD-WBR}(ours)} & \textbf{0.520} & \textbf{0.507} & \textbf{0.467} & 0.409 & 0.335 & \textbf{0.223} & \textbf{0.114} & \textbf{0.037} & \textbf{0.004} \\
    & & \multirow{1}{*}{2} & & 0.515 & 0.500 & 0.466 & \textbf{0.415} & \textbf{0.338} & 0.221 & 0.108 & 0.035 & \textbf{0.004} \\
    \midrule
    \multirow{2}{*}{\textbf{Wells}} & \multirow{2}{*}{\textbf{$\lambda_3$}} & \multirow{1}{*}{0.5} & \multirow{2}{*}{\emph{HSD-WBR}(ours)} & 0.824 & 0.824 & 0.824 & 0.824 & 0.824 & 0.824 & 0.824 & 0.810 & 0.781 \\
    & & \multirow{1}{*}{2} & & \textbf{0.824} & \textbf{0.824} & \textbf{0.824} & \textbf{0.824} & \textbf{0.824} & \textbf{0.824} & \textbf{0.824} & \textbf{0.824} & \textbf{0.824} \\
  \bottomrule
  \multicolumn{13}{p{350pt}}{Being the only architecture using $\lambda_3$, only HSD-WBR is considered.}
  \end{tabular}
  }
\end{table}

\begin{table}[h!]
  \caption{VACVPlaque \emph{JTPR} (Equations \ref{eq:jtp_rate_inner_vacv}-\ref{eq:jtp_rate_outer_vacv}) at different $\lambda_1$.}
  \label{tab:vacv_ablation_lambda1_joint_tp_rate_performance}
  \centering
  \resizebox{0.75\linewidth}{!}{
  \begin{tabular}{ccccc}
   \toprule
    & & & inner nested & outer nested \\
    $\lambda$ & Value & Architecture & object $\uparrow$ & object $\uparrow$ \\
    & & & (\emph{Plaques}) & (\emph{Wells}) \\
    \midrule
    \multirow{6}{*}{\textbf{$\lambda_1$}} & \multirow{3}{*}{0.1} & \emph{HSD-WBR}(ours) & \underline{0.794} & \textbf{0.849}  \\
    & & \emph{HSD}(ours) & 0.773 & \underline{0.841} \\
    & & \emph{SD} (RI)\citep{Stardist18} & \textbf{0.863} & \textbf{0.849} \\
    \cmidrule(r){2-5}
    & \multirow{3}{*}{0.4} & \emph{HSD-WBR}(ours) & 0.784 & \underline{0.849} \\
    & & \emph{HSD}(ours) & \underline{0.796} & \textbf{0.857} \\
    & & \emph{SD} (RI)\citep{Stardist18} & \textbf{0.833} & \underline{0.849} \\
  \bottomrule
  \multicolumn{5}{p{300pt}}{\emph{SD} (FT) experiments were omitted since they were pretrained with default $\lambda$s. For those experiments please see tables in the main text.}
  \end{tabular}
  }
\end{table}

\begin{table}[h!]
  \caption{VACVPlaque \emph{JTPR} (Equations \ref{eq:jtp_rate_inner_vacv}-\ref{eq:jtp_rate_outer_vacv}) at different $\lambda_2$.}
  \label{tab:vacv_ablation_lambda2_joint_tp_rate_performance}
  \centering
  \resizebox{0.75\linewidth}{!}{
  \begin{tabular}{ccccc}
   \toprule
    & & & inner nested & outer nested \\
    $\lambda$ & Value & Architecture & object $\uparrow$ & object $\uparrow$ \\
    & & & (\emph{Plaques}) & (\emph{Wells}) \\
    \midrule
    \multirow{6}{*}{\textbf{$\lambda_2$}} & \multirow{3}{*}{5e-5} & \emph{HSD-WBR}(ours) & 0.771 & 0.833 \\
    & & \emph{HSD}(ours) & \underline{0.815} & \textbf{0.849} \\
    & & \emph{SD} (RI)\citep{Stardist18} & \textbf{0.871} & \textbf{0.849} \\
    \cmidrule(r){2-5}
    & \multirow{3}{*}{2e-4} & \emph{HSD-WBR}(ours) & \underline{0.793} & \underline{0.841} \\
    & & \emph{HSD}(ours) & 0.748 & 0.833 \\
    & & \emph{SD} (RI)\citep{Stardist18} & \textbf{0.858} & \textbf{0.849} \\
  \bottomrule
  \multicolumn{5}{p{300pt}}{\emph{SD} (FT) experiments were omitted since they were pretrained with default $\lambda$s. For those experiments please see tables in the main text.}
  \end{tabular}
  }
\end{table}

\begin{table}[h!]
  \caption{VACVPlaque \emph{JTPR} (Equations \ref{eq:jtp_rate_inner_vacv}-\ref{eq:jtp_rate_outer_vacv}) at different $\lambda_3$.}
  \label{tab:vacv_ablation_lambda3_joint_tp_rate_performance}
  \centering
  \resizebox{0.75\linewidth}{!}{
  \begin{tabular}{ccccc}
   \toprule
    & & & inner nested & outer nested \\
    $\lambda$ & Value & Architecture & object $\uparrow$ & object $\uparrow$ \\
    & & & (\emph{Plaques}) & (\emph{Wells}) \\
    \midrule
    \multirow{2}{*}{\textbf{$\lambda_3$}} & \multirow{1}{*}{0.5} & \multirow{2}{*}{\emph{HSD-WBR}(ours)} & \textbf{0.768} & \textbf{0.841} \\
    & 2 & & 0.721 & 0.825 \\
  \bottomrule
  \multicolumn{5}{p{300pt}}{Being the only architecture using $\lambda_3$, only HSD-WBR is considered.}
  \end{tabular}
  }
\end{table}

\end{document}